\RequirePackage{fix-cm}
\documentclass[twocolumn]{svjour3}          
\smartqed  
\usepackage{microtype}
\usepackage{graphicx}
\usepackage{subfigure}
\usepackage{booktabs} 
\usepackage{hyperref}

\usepackage{natbib}

\usepackage[utf8]{inputenc} 
\usepackage[T1]{fontenc}    
\usepackage{hyperref}       
\usepackage{url}            
\usepackage{booktabs}       
\usepackage{amsfonts}       
\usepackage{nicefrac}       
\usepackage{microtype}      
\usepackage{graphicx}
\usepackage{wrapfig}

\usepackage{amsmath}
\usepackage{amssymb}
\usepackage{algorithm} 
\usepackage{algorithmic} 
\usepackage{enumerate} 
\usepackage{multirow} 
\usepackage{amsmath}

\usepackage{subfigure}

\usepackage{helvet} 
\usepackage{courier} 
\usepackage{xcolor}

\newcommand{\Umat}[0]{\ensuremath{{\bf U}} }

\newcommand{\Wmat}[0]{\ensuremath{{\bf W}} }
\newcommand{\Xmat}[0]{\ensuremath{{\bf X}} }

\newcommand{\av}[0]{\ensuremath{\boldsymbol{a}} }
\newcommand{\bv}[0]{\ensuremath{\boldsymbol{b}} }

\newcommand{\dv}[0]{\ensuremath{\boldsymbol{d}} }

\newcommand{\hv}[0]{\ensuremath{\boldsymbol{h}} }

\newcommand{\kv}[0]{\ensuremath{\boldsymbol{k}} }

\newcommand{\mv}[0]{\ensuremath{\boldsymbol{m}} }

\newcommand{\pv}[0]{\ensuremath{\boldsymbol{p}} }

\newcommand{\rv}[0]{\ensuremath{\boldsymbol{r}} }

\newcommand{\uv}[0]{\ensuremath{\boldsymbol{u}} }
\newcommand{\vv}[0]{\ensuremath{\boldsymbol{v}} }
\newcommand{\wv}[0]{\ensuremath{\boldsymbol{w}} }
\newcommand{\xv}[0]{\ensuremath{\boldsymbol{x}} }

\newcommand{\zv}[0]{\ensuremath{\boldsymbol{z}} }
\newcommand{\Av}[0]{\ensuremath{\boldsymbol{A}} }

\newcommand{\Phimat}[0]{\ensuremath{\boldsymbol{\Phi}}}

\newcommand{\Omegamat}[0]{\ensuremath{\boldsymbol{\Omega}}}

\newcommand{\epsilonv}[0]{\ensuremath{\boldsymbol{\epsilon}} }

\newcommand{\thetav}[0]{\ensuremath{\boldsymbol{\theta}} }

\newcommand{\lambdav}[0]{\ensuremath{\boldsymbol{\lambda}} }

\newcommand{\phiv}[0]{\ensuremath{\boldsymbol{\phi}} }

\newcommand{\cdotv}[0]{\ensuremath{\boldsymbol{\cdot}}}

\newcommand{\given}{\,|\,}

\begin{document}

\title{Matching Visual Features to Hierarchical Semantic Topics for \\ Image Paragraph Captioning
}


\author{Dandan Guo*  \and Ruiying Lu* \and Bo Chen \and Zequn Zeng \and Mingyuan Zhou}


\institute{Dandan Guo \and Ruiying Lu \and Bo Chen  \and Zequn Zeng \at
              National Laboratory of Radar Signal Processing, Collaborative Innovation Center of Information Sensing and Understanding, Xidian University, Xi'an 710071, China. \\
              \email{gdd\_xidian@126.com; ruiyinglu\_xidian@163.com;\\bchen@mail.xidian.edu.cn;
              zzq8341@gmail.com}           
           \and
           Mingyuan Zhou \at
            McCombs School of Business, The University of Texas at Austin, Austin, TX 78712, USA.\\
             \email{mingyuan.zhou@mccombs.utexas.edu}\\ *Equal contribution. \\Corresponding author: Bo     Chen.
}

\maketitle

\begin{abstract}
Observing a set of images and their corresponding paragraph-captions, a challenging task is to learn how to produce a semantically coherent paragraph to describe the visual content of an image. Inspired by recent successes in integrating semantic topics into this task, this paper develops a plug-and-play hierarchical-topic-guided image paragraph generation framework, which couples a visual extractor with a deep topic model to guide the learning of a language model. To capture the correlations between the image and text at multiple levels of abstraction and learn the semantic topics from images, we design a variational inference network to build the mapping from image features to textual captions. To guide the paragraph generation, the learned hierarchical topics and visual features are integrated into the language model, including Long Short-Term Memory (LSTM) and Transformer, and jointly optimized.
Experiments on public datasets demonstrate that the proposed models, which are competitive with many  state-of-the-art approaches in terms of standard evaluation metrics, can be used to both distill interpretable multi-layer semantic topics and generate diverse and coherent captions. We release our code at https://github.com/DandanGuo1993/VTCM-based-image-paragraph-caption.git
\end{abstract}

\section{Introduction}\label{intro}
Describing visual content in a natural-language utterance is an emerging interdisciplinary problem, which lies at the intersection of computer vision (CV) and natural language processing (NLP) (\citealt{ordonez2016large}). As a sentence-level short image caption  (\citealt{Showattend_tell,vinyals2015show,BottomUp_imagecaption}) has a limited descriptive capacity, \cite{Regions-Hierarchical} introduce a paragraph-level captioning method that aims to generate a detailed and coherent paragraph for describing an image more finely.
Recent advances in image paragraph generation focus on building different types of hierarchical recurrent neural network (HRNN), $e.g.,$ LSTM (\citealt{LSTM}), to generate the visual paragraphs. For HRNN, the high-level RNN takes the image features as the input and recursively produces a sequence of sentence-level vectors, which are often explained as the topic vectors; while the low-level RNN is subsequently adopted to decode each topic vector into an output sentence.
By modeling each sentence and coupling the sentences into one paragraph, these hierarchical architectures often outperform existing flat models (\citealt{Regions-Hierarchical}). To improve the performance and generate more diverse paragraphs, advanced methods, extending HRNN based on generative adversarial networks (GAN) (\citealt{GAN}) or variational auto-encoders (VAE) (\citealt{VAE}), are proposed by \cite{RTT-GAN} and \cite{CapG-RevG}. Apart from adopting the output of the high-level RNN to represent the topics, \cite{wang2019convolutional}
introduce convolutional auto-encoding (CAE) on the region-level features of an image to learn the corresponding topics, which are further integrated into the HRNN-based paragraph generation framework.

In summary, the above image paragraph captioning methods typically refer to the output of the high-level RNN or CAE as the topics. Note that these topics are very different from the semantic topics represented by a set of semantically related words, explored in  topic models (TMs). Designing topics in this way may cause these above models only attend to some visually salient image regions without grasping the image's main semantic topic. As discussed by \citealt{topic-guided}, having an intuition about an image's high-level semantic topics is generally beneficial for selecting the most semantically-meaningful and topic-relevant image areas for describing an image. Recently, there are several attempts to utilize semantic topics learned from topic models, $e.g.$, Latent Dirichlet Allocation (LDA) (\citealt{blei2003latent}), a commonly used shallow topic model, to generate a single-sentence caption (\citealt{fu2017aligning,topic-guided,yu2018topic,TOMS}). Similar to them, \cite{TOMS} extract textual topics of images with LDA, and generate topic-oriented multiple sentences to describe an image. Although having been proved effective, these above methods of utilizing semantic topics still have clear limitations. Typically, they only utilize shallow topic models to extract the single-layer semantic topics, which may have limited representation capacity. Another key limitation lies in adopting a two-stage manner to extract the semantic topics from images. Usually, they pre-train LDA from the captions of the training images and then train a downstream topic estimator to predict the topics with the image features as the input. However, this two-stage way does not consider the visual image features when learning the topic information and discard the uncertainty of the topic information brought by the probabilistic
topic model, which is a desired property to capture the inherent ambiguity of paragraph generation from images.

This paper presents a flexible hierarchical-topic-guided image paragraph generation framework in an end-to-end manner, coupling a visual extractor with a multi-stochastic-layer deep topic model to guide the generation of a language model (LM). Specifically, a convolutional neural network (CNN) coupled with a region proposal network (\citealt{Faster}) is first utilized to detect a set of salient image regions as the visual extractor, a usual practice in image captioning systems. Motivated by the idea of using multi-layer features in \cite{xu2015discriminative} and \citet{zhu2021temporal}, we aim to extract the multi-stochastic-layer topics and use them to guide the paragraph generation semantically, which has not been well exploited in existing methods for image paragraph captioning. To this end, we construct a deep topic model to match the image's visual features to its corresponding semantic topic information. Here we design a deep topic model built on the success of the Poisson gamma belief network (PGBN) (\citealt{GBN}), which extracts interpretable multi-layer topics from text data and can be equivalently represented as deep LDA (\citealt{cong2017deep}). A naive approach by introducing PGBN into the image paragraph caption task is adopting the two-stage manner similar to \cite{TOMS}, where we can pre-train PGBN, extract hierarchical topics from the training captions, and then build a downstream topic classifier with a deterministic network to approximate the deep topics from the image features. However, this inflexible way does not incorporate the visual image features into the learning process of the hierarchical topics and abandons the uncertainty in the probabilistic topic model (PGBN), resulting in unsatisfactory
image paragraph captioning performance. To capture the correlations between the image and text at multiple levels of abstraction and learn the semantic topics from images, we here generalize PGBN into a novel visual-textual coupling model (VTCM). {Generally,} VTCM encapsulates region-level features into the hierarchical topics by a variational encoder and feeds the topic usage information to the decoder (PGBN) to generate descriptive captions. %
Different from existing image paragraph caption methods that compute topic vectors via deterministic neural networks (RNN or CAE) or {learn the semantic topics using the shallow topic models (typically in a two-stage manner)}, our proposed VTCM can relate semantic topics and visual concepts and distill the multi-stochastic-layer topic information in an ``end-to-end'' manner.

To guide paragraph generation, both the visual features and mined hierarchical semantic topics from the VTCM are fed into either an LSTM or Transformer (\citealt{Transformer}) based language generator. We refer to them as VTCM-LSTM and VTCM-Transformer. Following \cite{wang2019convolutional}, the LM in VTCM-LSTM capitalizes on both paragraph-level and sentence-level LSTMs. Inspired by the idea of selecting top-relevant regions \citep{BottomUp_imagecaption, fan2020recurrent}, the feedback of the paragraph-level LSTM is fed into the attention module together with the topic information to select critical image visual features. The sentence-level LSTM generates a sequence of words conditioning on the learned topics and attended image features. For Transformer-based image caption systems, while the original Transformer architecture can be directly adopted as the LM in our framework, the multi-modal nature of image captions
requires specialized architectures different from those employed for the understanding of a single modality.
{\citet{cornia2020meshed} thus introduce a Meshed-Transformer with memory for image captioning, which learns a multi-level representation of the relationships between image regions via an augmented-memory encoder and uses mesh-like connectivity at the decoding stage to exploit both low- and high-level features. Our work aims to improve the Meshed-Transformer with the multi-stochastic-layer topic information, which is hierarchically coupled with the visual features extracted by the encoder and further interpolated into the feedback of each decoding layer to guide the caption generation.}
Absorbing the multi-layer semantic topics as additional guidance, both VTCM-LSTM and VTCM-Transformer produce a caption closely related to the given image and semantic topics.
Unlike previous works that adopt GAN or VAE to enforce diversity in the generated captions (\citealt{RTT-GAN,CapG-RevG}), our paragraph captioning systems can generate diverse captions for an image since we feed the multi-stochastic-layer latent topic representation of VTCM as the source of randomness to the language generator.
Moving beyond existing methods that often use pre-trained semantic topics learned from LDA (\citealt{fu2017aligning,TOMS,yu2018topic,chen2019topics}),  our model allows jointly training the proposed VTCM (a deep topic model) and LM in an end-to-end manner. Since the semantic topics learned with VTCM are represented with a set of keywords, we can designate different topics as high-level guiding information, where the generated captions can be not only related to the image but also reflect what the user wants to emphasize. To the best of our knowledge, we are the first to distill the hierarchical semantic topics by capturing the correlations between the image and text at multiple levels of abstraction, and feed the topics into an LSTM-based or Transformer-based LM in an end-to-end manner to guide the paragraph generation. Due to the effectiveness and flexibility of our proposed plug-and-play system, one can also replace the  language model with other architectures.
Our main contributions include: 
1) VTCM is proposed to extract and relate the hierarchical semantic topics with image features,
where the distilled topics are integrated into both LSTM-based and Transformer-based LMs, guiding paragraph-level caption generation; 2) An end-to-end training is introduced to optimize the VTCM and LM jointly, beneficial for relating the visual and semantic concepts; 3) Extensive experiments are performed, with the quantitative and qualitative results showing the benefits of extracting multi-layer semantic topics for generating descriptive paragraphs.
\section{Related Work}\label{sec:1}
Below we review related work on image paragraph captioning, topic molding, and language modeling.
\subsection{Image Paragraph Captioning}

Image captioning aims to describe images with natural language, in which a popular research line is generating a single sentence to depict an image (\citealt{vinyals2015show,Showattend_tell}), denoted as image sentence captioning. However, as a single-sentence description is often too short to capture all detailed information, image paragraph captioning has been proposed to describe an image by generating a paragraph consisting of multiple sentences. Besides, other research directions in image captioning have also gradually attracted attention. For example, visual storytelling (\citealt{tang2019show}) aims to generate narrative creations from ordered photo sequences. In addition, optical character recognition (OCR) based image captioning (\citealt{wang2021improving}) aims to automatically describe an image with a sentence according to all the visual entities (both visual objects and scene text) in the image. While these problems are also challenging and attractive, they are beyond the scope of this work that is focused on image paragraph captioning. Regions-Hierarchical (\citealt{Regions-Hierarchical}) designs HRNN to produce a generic paragraph for an image. To generate diverse and semantically coherent paragraphs, \cite{RTT-GAN} extend the HRNN by proposing an adversarial framework between structured paragraph generator and multi-level paragraph discriminators. Considering the difficulties associated with training GANs and deficiency of explicit coherence model, \cite{CapG-RevG} augment HRNN with coherence vectors and a formulation of VAE \citep{VAE}. To encapsulate region-level features of an image into the topics, \cite{wang2019convolutional} design a convolutional  auto-encoding (CAE) module for topic modeling, where the extracted topics are further integrated into a two-level LSTM-based paragraph generator. Motivated by some models that utilize semantic topics to generate single-sentence captions (\citealt{fu2017aligning,topic-guided,yu2018topic}), \cite{TOMS} pre-train the LDA (\citealt{blei2003latent}) from the caption corpus of the training images at the first step, then train a topic classifier for semantic regularization and topic prediction based on the learned topics. In short, most of these existing paragraph captioning models either adopt deterministic networks ($e.g.$, high-level RNN or CAE) to construct a topic for each sentence within the whole paragraph or utilize a pre-trained LDA in a two-stage manner to learn the shallow semantic topics of images.
Different from them, this work distills the multi-stochastic-layer semantic topics by capturing the correlations between the image and text at multiple levels of abstraction.

\subsection{Topic Models and Language Models}
Probabilistic topic models (PTMs), such as latent Dirichlet allocation (LDA) \citep{blei2003latent,griffiths2004finding} and Poisson factor analysis (PFA) \citep{zhou2012beta}, often represent each document as a bag of words (BoW), capturing global semantic coherency into semantically meaningful topics.
To explore the hierarchical semantic structures, PGBN (\citealt{GBN}), a deep generalization of PFA that can also be viewed as a deep LDA (\citealt{cong2017deep}), is proposed to extract interpretable hierarchical topics and capture the relationship between latent topics across multiple stochastic layers. Despite its effectiveness, PGBN, requiring the texts at training and testing stages, is not suitable for the image paragraph captioning task, where only images are given during the testing stage. To this end, we develop a variational encoder to match the visual features to the hierarchical topic information and generalize PGBN to build a novel visual-textual coupling model (VTCM), which can be jointly optimized with the paragraph generator. Although the idea of introducing a variational encoder is similar to neural topic models (NTMs) \citep{miao2016neural,ProdLDA,DVAE,zhang2018whai,zhao2021topic}, our VTCM captures the correlations between the image and text at multiple levels and is proposed to adapt to the image paragraph captioning task. To our knowledge, the works that connect deep topic modeling with visual features are still very limited. Different from \cite{zhang2020variational} that propose to match visual features with a topic model, where the topics are fed into a GAN-based image generator, we focus on integrating the learned hierarchical topics from VTCM into the LMs to guide the paragraph generation, a task distinct from image generation.

Existing LMs are often built on either recurrent units, as used in RNNs \citep{RNNencoder-decoder,LSTM}, or purely the attention mechanism based modules, as used in Transformer and its various generalizations \citep{Transformer,Gpt}. {RNN-based LMs have been successfully used in image paragraph captioning systems (discussed above), while single-sentence Transformer-based image captioning systems have started to attract attention \citep{huang2019attention,li2019entangled,cornia2020meshed}, not to mention the research on image paragraph captioning}. Note that we can flexibly select the LM for our plug-and-play system since we pay more attention to assimilating the multi-layer semantic topic information into the paragraph generator.
We consider both the LSTM-based and Transformer-based LMs to investigate the effectiveness of integrating the deep semantic topics.

\section{Proposed Models}
\begin{figure*}[!htbt]
\begin{center}
\includegraphics[height=6.4cm,width=14.2cm]{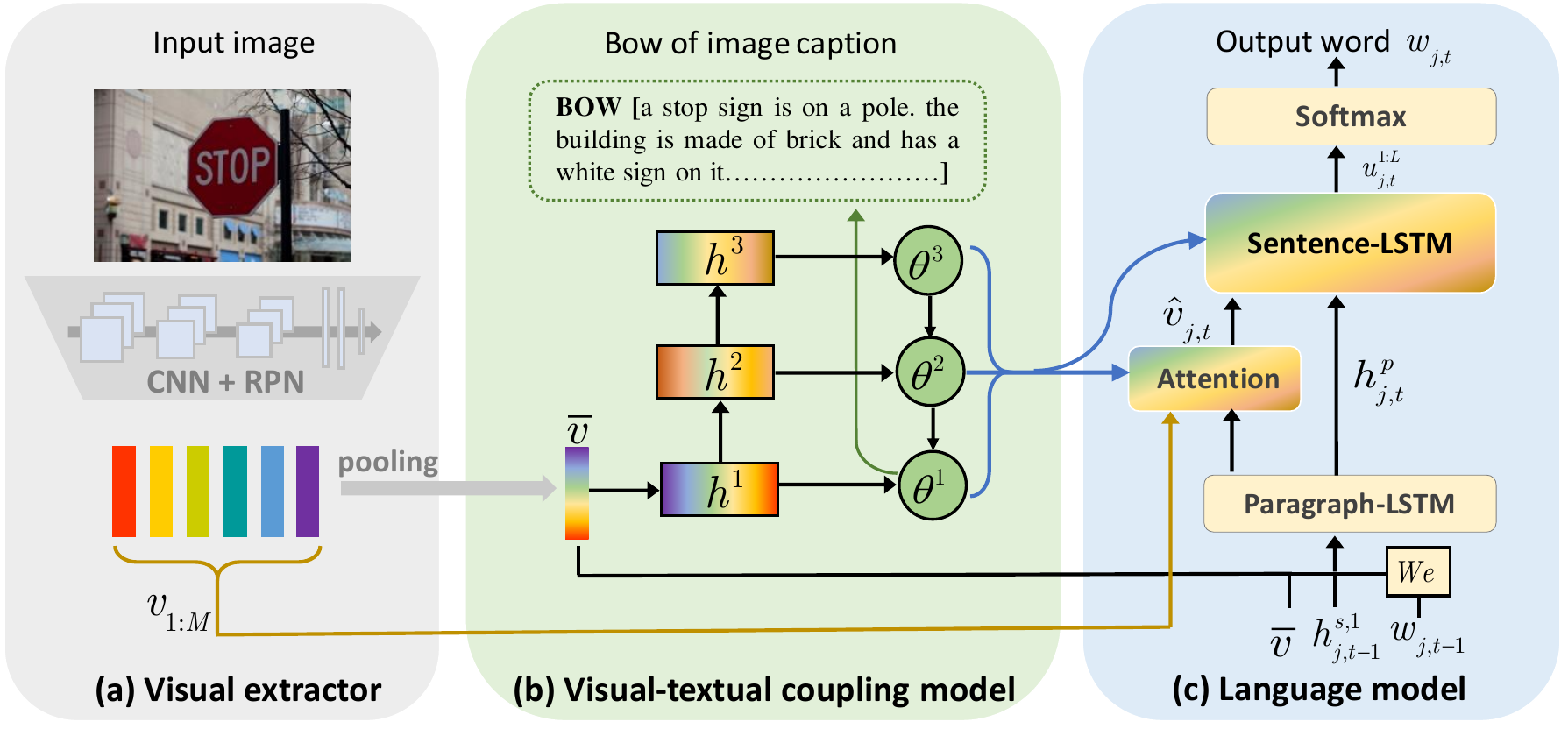}
\caption{{Architecture of our proposed VTCM-LSTM. (a) The visual extractor, consisting of a CNN and an RPN, produces  feature vectors $\vv_{1:M}$ and average-pooled vector $\overline{\vv}$. (b) The visual-textual coupling model, where the right part (from $\thetav^{3}$ to BoW of the paragraph caption) is the generative model with a three-hidden-layer (decoder) and left (from the average-pooled vector to the  $\hv^{l}$ and $\thetav^{l}$) is the variational encoder. (c) The LSTM-based LM, including paragraph-level LSTM, attention module and sentence-level LSTM, where $w_{j,t}$ is the $t$-th word in the $j$-th sentence of a paragraph and $\Wmat_e$ is the word embedding matrix.}}
\label{fig:overall_model}
\end{center}
\end{figure*}

Denoting $\emph{Img}$ as the given image, image paragraph captioning systems aim to generate a paragraph $P=\{S_1,...,S_J\}$ consisting of $J$ sentences, where sentence $S_j= \{w_{j,1},...,w_{j,T_j}\}$  consists of $T_j$ words from a vocabulary of size $V$. We introduce a plug-and-play hierarchical-topic-guided image paragraph captioning system, {with the overview of VTCM-LSTM depicted in Fig. \ref{fig:overall_model}.} It contains three major components, including a visual extractor for extracting image features, the proposed VTCM for distilling  multi-layer {semantic} topics of a given image, and the LM for interpreting the extracted image features and {semantic} topics into captions. Following a usual practice in image captioning systems, we implement the visual extractor by adopting a CNN coupled with a region proposal network (RPN) (\citealt{Faster}), as shown in Fig. {\ref{fig:overall_model}(a)}. The process is expressed as $\{\vv_{1},\cdots,\vv_{M}\}= \mbox{VE}(\emph{Img})$, where $\mbox{VE}(\cdot)$ denotes the visual extractor, $M$ the number of regions, $\vv_{i} \in \mathbb{R}^{D}$  the $i$-th salient region of \textit{Img} and $D$ the dimension of visual features. To further compactly describe the content of the image, we subsequently aggregate these $M$ vectors into a single average-pooled vector $\overline{\vv}=\frac{1}{M} \sum_{i} \vv_{i}$. Below, we give more details about the other two components, $i.e.,$ the VTCM and LM.

\subsection{Visual-Textual Coupling Model
}\label{topic model}
There are two mainstream ways to learn topics from an image. One is to encode the visual image features into a global vector with a high-level RNN, which is {explained as the topic vectors and used to guide a low-level RNN. Note that these topics are very different from the semantic topics, which are usually represented in the form of semantically related words in a topic model.} {Another way is first applying LDA  on the training captions and then training a downstream topic predictor over image features, where the semantic topics are assimilated into the LM for sentence or paragraph generation. Moving beyond them,} we design an {end-to-end} variational deep topic model to capture the correlations between image features and descriptive text by distilling the semantic topics, jointly trained with the LM. The basic idea follows the philosophy that the generation from topics to descriptive captions via topic decoder and the topics extraction over image visual features via variational encoder can enforce the mined multi-layer topics to be related to the visual features.

\noindent\textbf{\textrm{Topic Decoder}}:  As a multi-stochastic-layer deep generalization of LDA (\citealt{cong2017deep}), PGBN (\citealt{GBN}) is selected as the topic decoder. For the given $\emph{Img}$ in the training set, we summarize its ground-truth paragraph caption $P$ into a BoW count vector $\dv \in \mathbb{Z}_{+}^{V_c}$, where $V_c$ is the size of the vocabulary excluding stop words, ${Z}_{+}$ denotes non-negative integers, and each element of $\dv$ counts the number of times the corresponding word occurs in the paragraph. {As shown in Fig. {\ref{fig:overall_model}(b)}}, the generative process of PGBN with $L$-hidden-layer from top to bottom, is expressed as
\begin{align}\label{gbn}
\thetav^L \sim \textrm{Gamma}\left(\rv, \tau^{L+1}\right), \cdots,\notag \\
  \thetav^{l}\sim \textrm{Gamma}\left(\Phimat^{l+1} \thetav^{t+1}, \tau^{t+1}\right),  \cdots, \\
   \dv \sim \textrm{Poisson}(\Phimat^{1} \thetav^{1}),\thetav^{1} \sim \textrm{Gamma}\left(\Phimat^{2} \thetav^{2}, \tau^{2}\right), \notag
\end{align}
where the shape parameters of the gamma distributed hidden units $ \thetav^l \in \mathbb{R}_+^{K_l}$ are factorized into the product of connection weight matrix  $ \Phimat^{l+1} \in \mathbb{R}_+^{K_{l} \times K_{l+1}}$ and hidden units $ \thetav^{l+1}$ of the next layer, $K_l$ is the number of topics at layer $l$ and $K_0=V_c$. The vector $\rv=\{r_1,...,r_{K_{L}}\}$ at the top layer denotes the gamma shape parameter of $\thetav^L$; and $\{\tau^{l}\}_{l=1}^{L}$ are gamma scale parameters. We place a Dirichlet prior on each column of $ \Phimat^{l}$ at each layer, denoted as $ \phiv^{l} \sim \textrm{Dir}(\eta^{l},...,\eta^{l})$, where $\eta^{l}$ is the prior of $ \phiv^{l}$. The global semantics of image captions in training dataset are compressed into {$ \Phimat^{1:L}$}, where $ \Phimat^{l} \in \mathbb{R}_+^{K_{l} \times K_{l+1}}$ denotes $K_{l+1}$ topics at layer $l$ and each column of $ \Phimat^{l}$ corresponds to a topic. To visualize the topic $\phiv_k^{l}$ at hidden layer $l$, we can map it to the $V_c$-dimensional observation space, expressed as $\left[ \prod_{p=1}^{l-1} \Phimat^{p} \right] \phiv_k^{l} \in \mathbb{R}_+^{V_c}$, which is a distribution over all words in the vocabulary. $\thetav^{1:L}$ denote the hierarchical topic proportions of BoW count vector $\dv$ over semantic topics $\Phimat^{1:L}$ and thus capture the semantic information of different levels about the given $\emph{Img}$. Therefore, we can build a better paragraph generator by integrating hierarchical topic weight vectors into the language model.

\noindent\textbf{\textrm{Variational Topic Encoder}}: \label{topic_encoder}
{Under the hierarchical generative model of PGBN, conditioned on the captions of the training images, the inference task here is to find the global hierarchical topis $ \Phimat^{1:L}$  (shared by all captions) and deep topic proportions $ \thetav^{1:L}$ (specific for each caption). To infer $ \Phimat^{1:L}$, we can adopt the topic-layer-adaptive stochastic gradient Riemannian Markov chain Monte Carlo (TLASGR-MCMC) developed by \citet{cong2017deep} to provide a scalable distributional estimate. Given the topics $ \Phimat^{1:L}$, inferring the topic proportions $ \thetav^{1:L}$ from descriptive caption like the typical PGBN does is however not suitable for the image captioning task, where only images are given during the testing stage. To this end, a naive approach is adopting the two-stage manner. Namely, we can learn deep topics $ \Phimat^{1:L}$ and topic proportions $\thetav^{1:L}$ using training captions and then learn a downstream topic predictor $f(\thetav^{1:L}|\overline{\vv})$ to approximate $\thetav^{1:L}$ by taking visual features $ \overline{\vv}$ as the input. But, this two-stage way ignores the visual image features when learning the topic information and discards the uncertainty brought by the probabilistic topic model, a key property to capture the inherent ambiguity of paragraph generation from images. Therefore, motivated by the variational hetero-encoder in \citet{zhang2020variational}, we develop a variational topic encoder to match the visual features $ \overline{\vv}$ to the hierarchical topic weight vectors $ \thetav^{1:L}$ and generalize PGBN into a novel visual-textual coupling model (VTCM). Specifically, we build a topic encoder as $ \prod_{l=1}^{L} q(\thetav^{l}\given \overline{\vv} )$, with}
\begin{equation}\label{theta_Weibull}
  q(\thetav^{l}\given \overline{\vv}) = \mbox{Weibull}(\kv^{l},\lambdav^{l}),
\end{equation}
where the Weibull distribution is used to approximate the gamma distributed conditional posterior, and its corresponding parameters
$\kv^{l}, \lambdav^{l} \in \mathbb{R}_+^{K_l}$ are {deterministically nonlinearly transformed from the representation
$\hv^{l}$, which is mapped from the image features $\overline{\vv}$,} as described in {Appendix \ref{sec:topic encoder}} and shown in Fig. {\ref{fig:overall_model}(b)}. {Using the reparameterization trick, we can sample the Weibull distributed topic weight vector $\thetav^{l}$ as}
\begin{align}\label{sample_theta}
 \thetav^{l} &  \!=\! {\lambdav^{l}} \left(-\ln(1-{\epsilonv^{l}})\right) ^ {1/{\kv^{l}}},~
 {\epsilonv^{l}}\!\sim \!\textstyle \prod_{k=1}^{K_{l}}\mbox{Uniform}(0,1).
\end{align}
{Benefiting from the variational framework, we can randomly draw $\thetav^{l}$ from the same latent space parameterized by $\kv^{l}, \lambdav^{l}$, where  different $\thetav^{l}$ can capture the inherent ambiguity ($i.e.$, diversity ) for the given image but have the same semantic information.} We denote $ \Omegamat_\text{TM}$ as the set of encoder parameters, which can be updated via stochastic gradient descent (SGD) by maximizing a lower bound of the log marginal likelihood of caption $\dv$ in \eqref{gbn}, formulated as
\begin{equation}\label{TM_ELBO}
{
\begin{aligned}
&L_\text{TM} \!=\!  \mathbb{E}_{q(\thetav^{1}\given \overline{\vv})} \left[ \ln p\left( \dv \given \, \Phimat^{1}\thetav^{1} \right)\right]-\\
&\sum_{l=1}^L \mathbb{E}_{q(\thetav^{l}\given\overline{\vv})} \left[ \ln q\left( \thetav^{l}\given \overline{\vv}  \right) -\ln p \left( \thetav^{l} \given
\Phimat^{l + 1}\thetav^{l+ 1}\right) \right].
\end{aligned}}
\end{equation}
Optimizing the above lower bound will encourage the multi-stochastic-layer topic weight vectors $ \thetav^{1:L}$  to capture holistic and representative information from the image and its corresponding caption. Serving as the bridge between two modalities, the hierarchical semantic topics can be further utilized to guide the caption generation of LMs. Note that we can flexibly select the LM for our plug-and-play system since we pay more attention to assimilating the multi-layer semantic topic weight vectors into the paragraph generator. Below we investigate how to integrate the topic information into not only LSTM-based but also Transformer-based LMs.
\\
\subsection{LSTM-based Language Generation Model}\label{RNN-based LM}
Inspired by \citet{wang2019convolutional}, who integrate the topics learned from CAE into a two-level LSTM-based
paragraph generation framework with the attention mechanism in \cite{BottomUp_imagecaption}, we design a paragraph generator with a hierarchy constructed by a paragraph-level LSTM, a sentence-level LSTM, and an attention module, shown in Fig.~\ref{fig:overall_model}(c). The paragraph-level LSTM first encodes the semantic regions based on all previous words into the paragraph state. Then the attention module selects semantic regions with the guidance of the current paragraph state and semantic topics of the image. Finally, the sentence-level LSTM incorporates the topics, attended image features, and current paragraph state to facilitate word generation.
\\
 \textbf{\textrm{Paragraph-level LSTM}}:
To generate $w_{j,t}$ as the $t$-th word of the $j$-th sentence in a paragraph caption, we set $\xv_{j,t}^{p}$ as the input vector of the paragraph-level LSTM. By concatenating the previous output $ \hv_{j,t-1}^{s,1} \in \mathbb{R}^{H}$ of the sentence-level LSTM at layer $1$, the image feature $\overline{\vv}$, and previously generated word $w_{j,t-1}$, the $ \xv_{j,t}^{p}$ is formulated as $ \xv_{j,t}^{p}=\left[\hv_{j,t-1}^{s,1}, \overline{\boldsymbol{v}}, \Wmat_{e} w_{j,t-1}\right]$, where $ [\cdot, \cdot]$ indicates concatenation, $ \Wmat_{e} \in \mathbb{R}^{E \times V} $ is a word embedding matrix, $V$ the vocabulary size of LM, $E$ the embedding size, and $H$ the size of hidden state unit. This input provides paragraph-level LSTM the maximum contextual information, capturing both visual semantics of the image and long-range inter-sentence dependency within a paragraph caption \citep{BottomUp_imagecaption}. Then the hidden state of the paragraph-level LSTM is computed~as
\begin{equation}
\hv_{j,t}^{p}=\mathrm{LSTM}_{\mathrm{para}}(\xv_{j,t}^{p},\hv_{j,t-1}^{p}),
\end{equation}
where $\hv_{j,t}^{p}\in \mathbb{R}^{H}$ and $ \hv_{j,0}^{p}=\hv_{j-1,T_{j-1}}^{p}$ are set to explore inter-sentence dependency.
\\
 \textbf{\textrm{Attention Module}}:  Given the paragraph state $ \hv_{j,t}^{p}$ and the concatenation of multi-layer topic weight vectors $\thetav^{1:L}$, denoted as  $ [\thetav^{1:L}] \in \mathbb{R}_+^{\sum_{l=1}^{L} K_l}$, we build an {attention} module to select the most
information-carrying regions of the visual features for predicting $w_{j,t}$, defined as
\begin{align}\label{atten_LSTM}
a_{j,t}^{m} &\!=\! \wv_{att} \tanh \left(\Wmat_{va} \vv_{m}\!+\!\Wmat_{ha} \hv_{j,t}^{p} \!+\! \Wmat_{ta} [\thetav^{1:L}] \right) ,\notag\\
\pv_{j,t} &=\operatorname{softmax}\left(\av_{j,t}\right),
\end{align}
where $a_{j,t}^{m}$ is the $m$-th element of $\av_{j,t} \in \mathbb{R}^{M} $, and $ \wv_{att} \in \mathbb{R}^{1 \times A} $, $ \Wmat_{va} \in \mathbb{R}^{A \times D} $, $ \Wmat_{ha} \in \mathbb{R}^{A \times H} $,  $ \Wmat_{ta} \in \mathbb{R}^{A \times (\sum_{l=1}^{L} K_{l})} $ are the learned parameters and $\operatorname{softmax}$ is a function that turns the $M$-dimensional vector into a non-negative vector with $M$ elements summed to $1$. Defined in this way $\pv_{j,t}$ is a probability vector over all regions in the image. The attended image feature is calculated with $\hat{\vv}_{j,t}=\sum_{m=1}^{M} p_{j,t}^{m} \vv_{m}$, providing a natural way to integrate the multi-layer {semantic} topics as auxiliary guidance when generating attention.
\\
\textbf{\textrm{Sentence-level LSTM}}:
The input vector $\xv_{j,t}^{s}$ to the sentence-level LSTM at each time step consists of the output $\hv_{j,t}^{p}$ of the paragraph-level LSTM, concatenated with the attended image feature $\hat{\vv}_{j,t}$, stated as $\xv_{j,t}^{s}=\left[\hat{\boldsymbol{v}}_{j,t},\hv_{j,t}^{p}\right]$. Specifically, sentence-level LSTM in turn produces a sequence of hidden states $\{\hv_{j,1}^{s,l},..,\hv_{j,T_{j}}^{s,l}\} \in \mathbb{R}^{H}$ at layer $l$, one for each sentence in the paragraph, denoted as
\begin{equation}\label{RNN_hiddenstate}
 \hv_{j,t}^{s,l}=\left\{\begin{array}{ll}{\mathrm{LSTM}_{\mathrm{sent}}^{l}\left(\hv_{j, t-1}^{s,l}, \xv_{j,t}^{s}\right)} ,& {\text { if } l=1 } ,\\
\mathrm{LSTM}_{\mathrm{sent}}^{l}\left(\hv_{j, t-1}^{s,l},\uv_{j,t}^{{l-1}} \right), & {\text { if } L\geq l>1},
\end{array}\right.
\end{equation}
where $\uv_{j,t}^{l}$ is the coupling vector combining the topic weight vectors $\thetav^{l}$ and hidden output of the sentence-level LSTM $\hv_{j,t}^{s,l}$ at each time step $t$.  Following \citet{guorecurrent}, we realize $\uv_{j,t}^{l}=g^{l}\left({\hv}_{j, t}^{s,l}, {\thetav}^{l}\right)$ with a gating unit similar to the gated recurrent unit (\citealt{RNNencoder-decoder}), described in Appendix \ref{sec:GRU_gate}. The probability over words in the dictionary can be predicted by taking a linear projection and a softmax operation over the concatenation of $\uv_{j,t}^{l}$ across all layers. This method enhances the representation power and, with skip connections from all hidden layers to the output \citep{alex2013}, mitigates the vanishing gradient problem. We denote the parameters of the LSTM-based LM as $\Omegamat_{\text{LSTM}}$.

\subsection{Transformer-based Language Generation Model}\label{Transformer-based}
{To demonstrate the proposed plug-and-play system}, we also explore how to integrate the multi-layer topic information into existing Transformer-based LMs, due to their representation power and computational efficiency coming from pure attention mechanisms. Typically, attention operates on a set of queries $ \boldsymbol{Q}$, keys $ \boldsymbol{K}$, and values $ \boldsymbol{V}$, defined as
$
\text { Attention }(\boldsymbol{Q}, \boldsymbol{K}, \boldsymbol{V})\!=\!\operatorname{softmax}\left({Q \boldsymbol{K}^{T}}/{\sqrt{d}}\right) \boldsymbol{V}
$, where $ \boldsymbol{Q}$ is a matrix of $n_{q}$ query vectors, both $ \boldsymbol{K}$ and $ \boldsymbol{V}$  contain $n_{k}$ keys and values, all with the same dimensionality, and $d$ is a scaling factor. \citet{cornia2020meshed} design a novel Transformer-based architecture to improve the image encoder and language decoder, and prove its effectiveness on sentence-level image captions. However, they neither consider the paragraph-level image captioning task nor the semantic topics underlying the image. On the basis of this Transformer-based architecture, we here devise a semantic topic-guided Transformer model, which is conceptually divided into an encoder and a decoder module, shown in {Fig. \ref{fig:transformer}}. Following \cite{cornia2020meshed}, the encoder processes region-level image features and devises the relationships between them. The decoder not only reads from the output of each encoding layer like that of \cite{cornia2020meshed} but also assimilates the hierarchical topic information to generate the paragraph caption word by word.
\\
\textbf{Memory-augmented Encoder}:
Denoting the aforementioned set of features $ \{\vv_{1},\cdots,\vv_{M}\}$
as $ \Xmat$ for clarity, we adopt the  memory-augmented attention operator to encode image regions and their relationships, defined as
\begin{gather}
\mathcal{M}_{\text {mem }}(\boldsymbol{X})=\text { Attention }\left(W_{q} \boldsymbol{X}, \boldsymbol{K}, \boldsymbol{V}\right),\notag\\
\boldsymbol{K} =\left[W_{k} \boldsymbol{X}, \boldsymbol{M}_{k}\right],~~
\boldsymbol{V} =\left[W_{v} \boldsymbol{X}, \boldsymbol{M}_{v}\right],
\end{gather}
where ${W_{q}, W_{k}, W_{v}}$ are matrices of learnable weights, a usual practice in the original Transformer, and $ \boldsymbol{M}_{k}$ and $\boldsymbol{M}_{v}$ are additional keys and values implemented as plain learnable memory matrices.
Following the implementation of \citet{Transformer}, the memory-augmented attention can be applied in a multi-head fashion, whose output can be fed into a feed-forward layer, denoted as $ \mathcal{F}(\cdot)$. Both the attention and feed-forward layers are encapsulated within a residual connection and a layer norm operation, denoted as $\text{AddNorm}(\cdot)$. For the encoder with $L$ layers, its $l$-th encoding layer is therefore defined as
\begin{gather}
\tilde{\boldsymbol{X}}^{l}\!=\!\text { AddNorm }\left(\mathcal{F}(\boldsymbol{Z}^{l})\right), \notag\\
\boldsymbol{Z}^{l}\!=\!\text { AddNorm }\!\left(\mathcal{M}_{\operatorname{mem}}(\tilde{\boldsymbol{X}}^{l-1})\!\right),
\end{gather}
where $ \tilde{\boldsymbol{X}}^{l} \in R^{d}$ and $ \tilde{\boldsymbol{X}}^{0} =\boldsymbol{X}$. A stack of $L$ encoding layers will produce a multilevel output $ \tilde{\mathcal{X}}=\left(\tilde{\boldsymbol{X}}^{1}, \ldots, \tilde{\boldsymbol{X}}^{L}\right)$.
\\
 {\textbf{Topic-guided Meshed Decoder}}:
Given the region encodings $ \tilde{\mathcal{X}}$ and topic information $\thetav^{1:L}$, our decoder {aims to generate} the paragraph caption, denoted as  $ \boldsymbol{Y}=\{y_1,...,y_I\}$ consisting of $I$ words for clarity. Inspired by \citet{cornia2020meshed}, we construct the {topic-guided} Meshed Attention operator to connect $ \boldsymbol{Y}$ to all elements in $ \tilde{\mathcal{X}}$ and $ \thetav^{1:L}$ {hierarchically} through gated cross-attentions, formulated as
\begin{gather}
{\mathcal{M}_{\text {mesh }}(\tilde{\mathcal{X}},\tilde{\thetav}^{1:L} , \boldsymbol{Y})= \sum_{l=1}^{L} \boldsymbol{\alpha}_{l} \odot \mathcal{C}\left(\tilde{\boldsymbol{X}}^{l}+\tilde{\thetav^{l}}, \boldsymbol{Y}\right)},
\end{gather}
where {we combine the topic information $\tilde{\thetav}^{l}$ and the hidden output of the memory-augmented encoder $\tilde{\boldsymbol{X}}^{l}$ at each layer $l$ although other choices are also available,} $ \tilde{\thetav^{l}} \in R^{d}$ is projected from $ \thetav^{l} \in R^{K_l}$ into the encoder embedding space, and ${\mathcal{C}(\cdot, \cdot)}$ stands for the cross-attention, computed using queries from the decoder and keys and values from the encoder and topic information:
 \begin{gather}
\mathcal{C}\left(\tilde{\boldsymbol{X}}^{l}\!+\!\tilde{\thetav^{l}}, \boldsymbol{Y}\right)\notag\\
=\text { Attention }\!\left(W_{q} \boldsymbol{Y}, W_{k} (\tilde{\boldsymbol{X}}^{l}\!+\!\tilde{\thetav^{l}}), W_{v} (\tilde{\boldsymbol{X}}^{l}\!+\!\tilde{\thetav^{l}})\!\right).
\end{gather}
By computing $\boldsymbol{\alpha}_{l}=\operatorname{sigmoid}\left(W_{l}\left[\boldsymbol{Y}, \mathcal{C}\left(\tilde{\boldsymbol{X}}^{l}\!+\!\tilde{\thetav^{l}}, \boldsymbol{Y}\right)\right]\right)
$, we can measure the relevance between cross-attention results, where $ W_{l}$ is the learned weight matrix. Similar to the encoding layer, the final structure of each decoding layer is written as
\begin{gather}
\tilde{\boldsymbol{Y}}^{l}=\operatorname{AddNorm}(\mathcal{F}(\boldsymbol{Z}^{l}+\tilde{\thetav}^{l})),\\
\small{\boldsymbol{Z}^{l}\!=\!\operatorname{AddNorm}\left(\mathcal{M}_{\operatorname{mesh}}\!\left(\tilde{\mathcal{X}}\!,
\tilde{\thetav}^{1:L}\!,\! \text { AddNorm}\!\left(\mathcal{S}_{\operatorname{m}}\!(\tilde{\boldsymbol{Y}}^{l-1}\!)\!\right)\!\right)\right),}  \notag
\end{gather}
where $ \mathcal{S}_{\operatorname{m}}$ is a masked self-attention used in the original Transformer (\citealt{Transformer}), due to the prediction of a word should only depend on previously predicted words, and ${\tilde{\boldsymbol{Y}}^{0}=\boldsymbol{Y}}$. After taking a linear projection and a softmax operation over $\tilde{\boldsymbol{Y}}^{L}$, our decoder finally predicts the probability over words in the vocabulary. Similar to the LSTM-based LM, Transformer-based LM is also guided by the {multi-layer semantic topics} and attended image features when generating the caption, whose parameters are represented as $ \Omegamat_{\text{Trans}}$.

\begin{figure}[!t]
\begin{center}
\includegraphics[height=2.9cm,width=8.2cm]{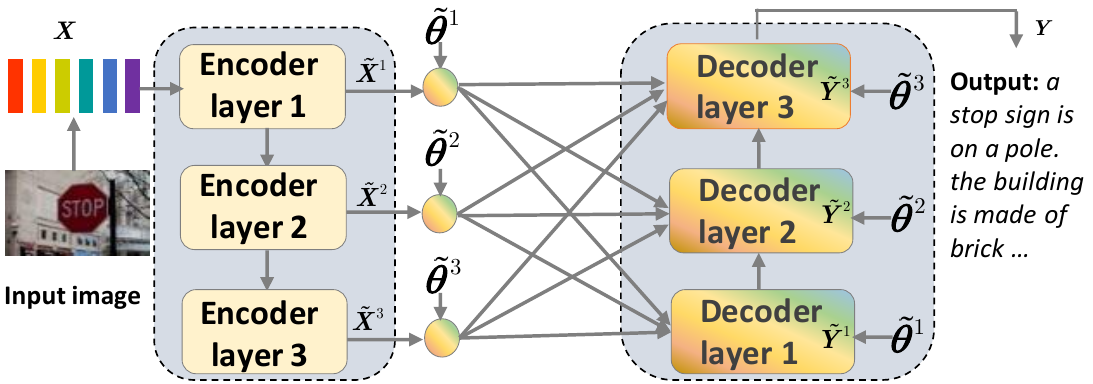}
\caption{{The overview of VTCM-Transformer, where VTCM is the same topic model used in VTCM-LSTM and omitted here.}}
\label{fig:transformer}
\end{center}
\end{figure}

\subsection{Joint Learning}\label{infer}
Under the deep topic model described in Section \ref{topic model} and LSTM-based LM in  Section \ref{RNN-based LM}, the joint likelihood of the target ground truth paragraph $P$ of $\emph{Img}$ and its corresponding BoW count vector $\dv$ is defined as
\begin{align}\label{likelihood2}
&\resizebox{0.99\hsize}{!}{$ p\left( P,\dv \given \Phimat^{1:L},\vv_{1:M}\right)\!=\!
 \int p\left(\dv\given \Phimat^{1}\thetav^{1}\right) \left[\prod_{l = 1}^{L} p\left(\thetav^{l}\given \Phimat^{l+1}\thetav^{l+1}\right) \right] $} \notag \\
&\resizebox{0.85\hsize}{!}{$ \prod_{j=1}^{J} \prod_{t=1}^{T_j}    p\left(w_{j,t} \given w_{j,<t}, \vv_{1:M},\thetav^{1:L}\right)
d{\thetav^{1:L}}$},
\end{align}
which is similar as the likelihood of the topic-guided Transformer-based captioning system, described in {Appendix \ref{sec:likelihood_VTCM_Transformer}}. As discussed in Section \ref{topic model}, we introduce a variational topic encoder to learn the multi-layer topic weight vectors $\thetav^{1:L}$ in \eqref{theta_Weibull} with the image features as the input. Thus,
a lower bound of the log of \eqref{likelihood2} can be constructed as
 \begin{gather}\label{ELBO-of-our model2}
 L_\text{all} = \mathbb{E}_{q(\thetav^{1}|\overline{\vv})}\left[ \ln p\left( \dv \given \, \Phimat^{1}\thetav^{1} \right)\right] \notag \\
-\sum_{l=1}^L \mathbb{E}_{q(\thetav^{l}|\overline{\vv})} \left[ \ln{q\left( \thetav^{l}\given \overline{\vv}  \right) }-\ln{p \left( \thetav^{l} \given \Phimat^{l + 1}\thetav^{l+ 1}\right)} \right] \\
+\sum_{l=1}^L \mathbb{E}_{q(\thetav^{l}|\overline{\vv})} \left[ \sum_{j=1}^{J}\sum_{t=1}^{T_j} \ln p\left( {w_{j,t}}\given w_{j,<t},\vv_{1:M},\thetav^{1:L}\right)\right],\notag
\end{gather}
which unites the first two terms primarily responsible for training the topic model component, and the last term for training the LM component. The parameters $\Omegamat_{\text{TM}}$ of the variational topic encoder and the parameters $\Omegamat_\text{LSTM}$ of LSTM-based LM can be jointly updated by maximizing  $L_\text{all}$. Besides, the global parameters $\Phimat^{1:L}$ of the topic decoder can be sampled with TLASGR-MCMC in \citet{cong2017deep}, described in Appendix~\ref{sec:update_Phi}. The training strategy is outlined in  Algorithm \ref{Algorithm}.

To sum up, as shown in Fig. \ref{fig:overall_model}, the proposed framework couples the topic model (VTCM) with a visual extractor, which takes the visual features of the given image as input and maps the hierarchical topic weight vectors.
The learned topic vectors in different layers are then used to reconstruct the BoW vector of the given image paragraph caption and as the additional features for the LSTM (or Transformer)-based LM to generate the paragraph.
Moreover, our proposed model introduces randomness into the topic weight vector, which captures the uncertainty about what is depicted in an image and hence encourages the diversity of generation.

\begin{algorithm}[!t]
\caption{ {Inference for our proposed VTCM-LSTM.}}
\begin{algorithmic}
 \STATE Set mini-batch size $ N$, the number of layer $ L$ and the width of layer $ K_l$; Initialize topic encoder parameters $ \Omegamat_\text{TM}$ and LSTM-based parameters $\Omegamat_\text{LSTM}$ and topic decoder parameters $ \Phimat^{1:L}$.
 \FOR{$iter = 1,2, \cdots$ }
 \STATE
 Randomly select a mini-batch of $N$ images and their paragraph captions to form a subset ${ \{ \emph{Img}_n, P_n, \dv_n \}_{n=1}^{N}}$.\\
 Compute the image features with visual extractor;\\
 Draw random noise $\{ {{\epsilonv _{n}^{l}}} \}_{n=1,l=1}^{N,L}$ from uniform distribution and sample latent states $\{\thetav_{n}^{ l }\}_{n = 1,l=1}^{N,L}$ from \eqref{sample_theta} via $\Omegamat_\text{TM}$, which are fed into the LSTM with \eqref{atten_LSTM} and \eqref{RNN_hiddenstate};  \\
 Compute $  {\nabla_{\Omegamat_\text{TM}}} L_\text{all}$ and $ {\nabla_{\Omegamat_\text{LSTM}}} L_{all} $ according to \ref{infer}, and update $\Omegamat_\text{TM}$ and $\Omegamat_\text{LSTM}$;\\
 \STATE Update $\Phimat^{1:L}$ with $\{\thetav_{n}^{ l }\}_{n = 1,l=1}^{N,L}$, described in  Appendix \ref{sec:update_Phi};\\
 \ENDFOR\\
\end{algorithmic}\label{Algorithm}
\end{algorithm}

\section{Experiments}
\subsection{Dataset and Implementation Details}
We conduct experiments on the public Stanford image-paragraph dataset (\citealt{Regions-Hierarchical}),
where 14,575 image-paragraph pairs are used for training, 2,487 for validation, and 2,489 for testing. Following the standard evaluation protocol, we use the full set of captioning metrics: METEOR (\citealt{Meteor}), CIDEr (\citealt{CIDEr}), and BLEU (\citealt{Bleu}). Different from the {BLEU} scores primarily measuring the $n$-gram precision, METEOR and CIDEr are known to provide more robust evaluations of language generation algorithms (\citealt{CIDEr}). In our experiments, the hyper-parameters and model checkpoints are chosen by optimizing the performance based on the average of {METEOR} and {CIDEr} scores on the validation set.
\\
{Following the publicly available implementation of \cite{BottomUp_imagecaption} and \cite{wang2019convolutional}, we use Faster R-CNN (\citealt{Faster}) with VGG16 network (\citealt{VGGnet}) as the visual extractor, which is pre-trained over Visual Genome (\citealt{krishna2017visual}). The top $M =50$ detected regions are selected to represent image features.}
The size of each image feature vector is $4096$, which is embedded into the $1024$-dimensional vector before being fed into our topic model. For our LMs, we tokenize words and sentences using Stanford CoreNLP (\citealt{CoreNLP}), lowercase all words, and filter out words that occur less than $1$ time. We set the maximum number of sentences in each paragraph as $6$ and the maximum length of each sentence as $30$ (padded where necessary) {for VTCM-LSTM}. For our topic model, all the words
from the training dataset, excluding the stopwords and the top $0.1\%$ most frequent words, are used to obtain a BoW caption for the corresponding image. The hidden sizes of paragraph-LSTM, sentence-LSTM, and attention module are all set to $512$. For our Transformer-based LM, we set the dimensionality $d$ of each layer {as} $512$, the number of heads {as} $8$, and the number of memory vectors as $40$. Both our Transformer-based and LSTM-based LMs are a three-layer model, same with the topic model with the topic number of $[K_1;K_2;K_3] = [80; 50; 30]$. Besides, we directly set hyper-parameters in VTCM as
$\{\eta^{l} = 0.1, \tau^{l}= 1, \rv = 1\}$. We use the Adam optimizer (\citealt{Adam}) with a learning rate of $5e-4$ for VTCM-LSTM and $1$ for VTCM-Transformer. The gradients of both VTCM-LSTM and VTCM-Transformer are clipped if the norm of the parameter vector exceeds $0.1$. The dropout rate is set to $0.5$ for VTCM-LSTM and 0.9 for VTCM-Transformer, and adopted in both the input and output layers to avoid overfitting. During inference, we adopt the penalty on trigram repetition proposed by \cite{TDIPC} and set the penalty hyperparameter as $2$. We also provide additional experimental results on the radiology report generation task in the Appendix~\ref{sec:additional_results}.
\\
\subsection{Baselines}
For {a} fair comparison, we consider the following baselines: 1) {Image-Flat}  (\citealt{Image-Flat}), directly decoding
a paragraph word-by-word via a single LSTM; 2) {Flat-repetition-penalty (\citealt{TDIPC}), training the non-hierarchical LSTM-based LM with an integrated penalty on trigram repetition to improve the diversity in image paragraph captioning;} 3) {Regions-Hierarchical} (\citealt{Regions-Hierarchical}), using a hierarchical LSTM to generate a paragraph, sentence by sentence; 4) {RTT-GAN} (\citealt{RTT-GAN}), training the Regions-Hierarchical in a GAN setting, coupled with an attention mechanism; 5) {TOMS} (\citealt{TOMS}), generating multi-sentences under the topic guidence, which trains a downstream topic classifier to predict the topics mined  by the LDA; 6) {Diverse-VAE} (\citealt{CapG-RevG}), leveraging coherence vectors and global topic vectors to generate paragraph, under a VAE framework; 7) IMAP (\citealt{xu2020interactive}), proposing an interactive key-value memory-augmented attention into the hierarchical LSTM; 8) {LSTM-ATT, which refers the outputs of paragraph-level LSTM as the topic vectors and adopts the attention mechanism in \cite{BottomUp_imagecaption}, as a degraded version of the proposed VTCM-LSTM;} 9) $\mathcal{M}^{2}$-Transformer (\citealt{cornia2020meshed}),  a novel Transformer-based architecture for single-sentence image captioning and a degraded version of VTCM-Transformer;
{10)  CAE-LSTM (\citealt{wang2019convolutional}), which adopts the CAE to extract topics and integrates
them into the two-level LSTM-based paragraph generator;}
11) Splitting to Tree Decoder (S2TD) (\citealt{shi2021s2td}), which models the paragraph decoding process as a top-down binary tree expansion and consists of a split module, a score module, and a word-level LSTM; 12) Retrieval-enhanced adversarial training with dynamic memory-augmented attention for image paragraph captioning (RAMP) (\citealt{xu2021retrieval}), which makes full use of the R-best retrieved candidate captions and adopts the hierarchical LSTM as the paragraph generator.

\begin{table*}[!ht]
\centering
\caption{ Main results for generating paragraphs. Our models are compared with competing baselines along with six language metrics. The human performance is included for providing a better understanding of all metrics following \cite{Regions-Hierarchical}.}
\resizebox{1\textwidth}{!}{
\begin{tabular}{c|cccccc}
\toprule[1pt]
{Method} & METEOR & CIDEr &BLEU-1 &BLEU-2 &BLEU-3 &BLEU-4\\ \hline
Image-Flat (\citealt{Image-Flat}) &12.82 &11.06 &34.04&19.95&12.20&7.71\\
{Flat-repetition-penalty} (\citealt{TDIPC}) &15.17 & 22.68 & 35.68 & 22.40 &14.04 & 8.70\\
TOMS (\citealt{TOMS}) &18.6 &20.8&{43.1}&{25.8}&14.3&8.4\\
Regions-Hierarchical (\citealt{Regions-Hierarchical}) &15.95&13.52&41.90&24.11&14.23&8.69\\
RTT-GAN (\citealt{RTT-GAN}) &17.12 &16.87 &41.99 &24.86 &14.89 &9.03\\
Diverse-VAE (\citealt{CapG-RevG}) &\bf{18.62}&20.93&42.38&25.52&15.15&9.43\\
IMAP (\citealt{xu2020interactive}) &16.56&20.76&42.38&{25.87}&15.51&9.42\\
S2TD(\citealt{shi2021s2td})&17.00& 21.92&\bf{44.59}& \bf{26.06}& 14.93 &8.35\\
\hline
{LSTM-ATT } & 17.40& 20.11 &40.8 &24.75 &14.81 &8.95\\
\bf{Our VTCM-LSTM} &17.52&\bf{22.82}&42.80&25.50&\bf{15.69}&\bf{9.63}\\
\hline
$\mathcal{M}^{2}$ Transformer (\citealt{cornia2020meshed}) & 15.4 &16.1 &37.5 &22.3 &13.7 &8.4\\
\bf{Our VTCM-Transformer} &16.88 &\bf{26.15} &40.93 &25.51 &\bf{15.94} &\bf{9.96}\\
\hline
{Human}  &19.22 &28.55 &42.88 &25.68 &15.55 &9.66 \\ \bottomrule
\end{tabular}}\label{results}
\end{table*}
\subsection{Quantitative Evaluation}
\textbf{Main Results}:
The results of different models on the Stanford dataset are shown in Table \ref{results}, where we only report the results of different models trained with cross-entropy rather than self-critical sequence training to eliminate the influence of different training strategies. As it can be observed, our proposed VTCM-LSTM surpasses all the other LSTM-based captioning systems in terms of BLEU-4, BLEU-3, and CIDEr, while being competitive on BLEU-1, BLEU-2 and METEOR with the best performer. Moreover, on all metrics, {our proposed VTCM-LSTM and VTCM-Transformer improve their corresponding baselines, $i.e.$, LSTM-ATT and $\mathcal{M}^{2}$-Transformer,  respectively.} These results demonstrate the effectiveness of integrating the semantic topics mined from VTCM into language generation in terms of topical semantics and descriptive completeness. {Moreover, VTCM-Transformer leads to a performance
boost over VTCM-LSTM on almost all the metrics, indicating the advantage of the memory-augmented operator and meshed cross-attention operator with a Transformer-like layer. We also replace the $\mathcal{M}^{2}$-Transformer with the original Transformer pretrained on a diverse set of unlabeled text, which however produces  poor performance, suggesting the importance of designing specialized architectures for multi-model image captioning.} Of particular note is the large improvement under both VTCM-LSTM and VTCM-Transformer on CIDEr, which is proposed specifically for image descriptions evaluation and measures the $n$-gram
accuracy by term-frequency inverse-document-frequency
(TF-IDF).  Interestingly, by bridging the visual features to the textual descriptions, our proposed VTCM is suited for extracting paragraph-level word concurrence patterns into latent topics, which capture the main aspects of the scene and image descriptions. The assimilation of topic information into language models thus leads to a large improvement in CIDEr, correlated well with human judgment. {However, it} is often not the case in other {image captioning systems} unless the CIDEr score is treated as the reward and directly optimized with policy-gradient based reinforcement learning techniques to finetune the model (\citealt{wang2019convolutional,cornia2020meshed,TDIPC,xu2020interactive}).

Notably, the CAE-LSTM of \citet{wang2019convolutional}, the S2TD of \citet{shi2021s2td}, and the RAMP of \citet{xu2021retrieval} additionally adopt self-critical (SC) training after the pre-training with cross-entropy (CE). For a fair comparison, we also treat the CIDEr as the reward and introduce the self-critical into the pre-trained VTCM-LSTM, where the results are reported in {Table \ref{RL_VTCM_LSTM}}. Our proposed model outperforms all the baselines on METEOR and CIDEr and achieves a comparable performance with S2TD and RAMP on BLEU-4. It indicates the effectiveness of assimilating the hierarchical semantic topic from VTCM into the paragraph generator.

\begin{figure*}[!htbt]
\begin{center}
\includegraphics[height=12cm,width=17cm]{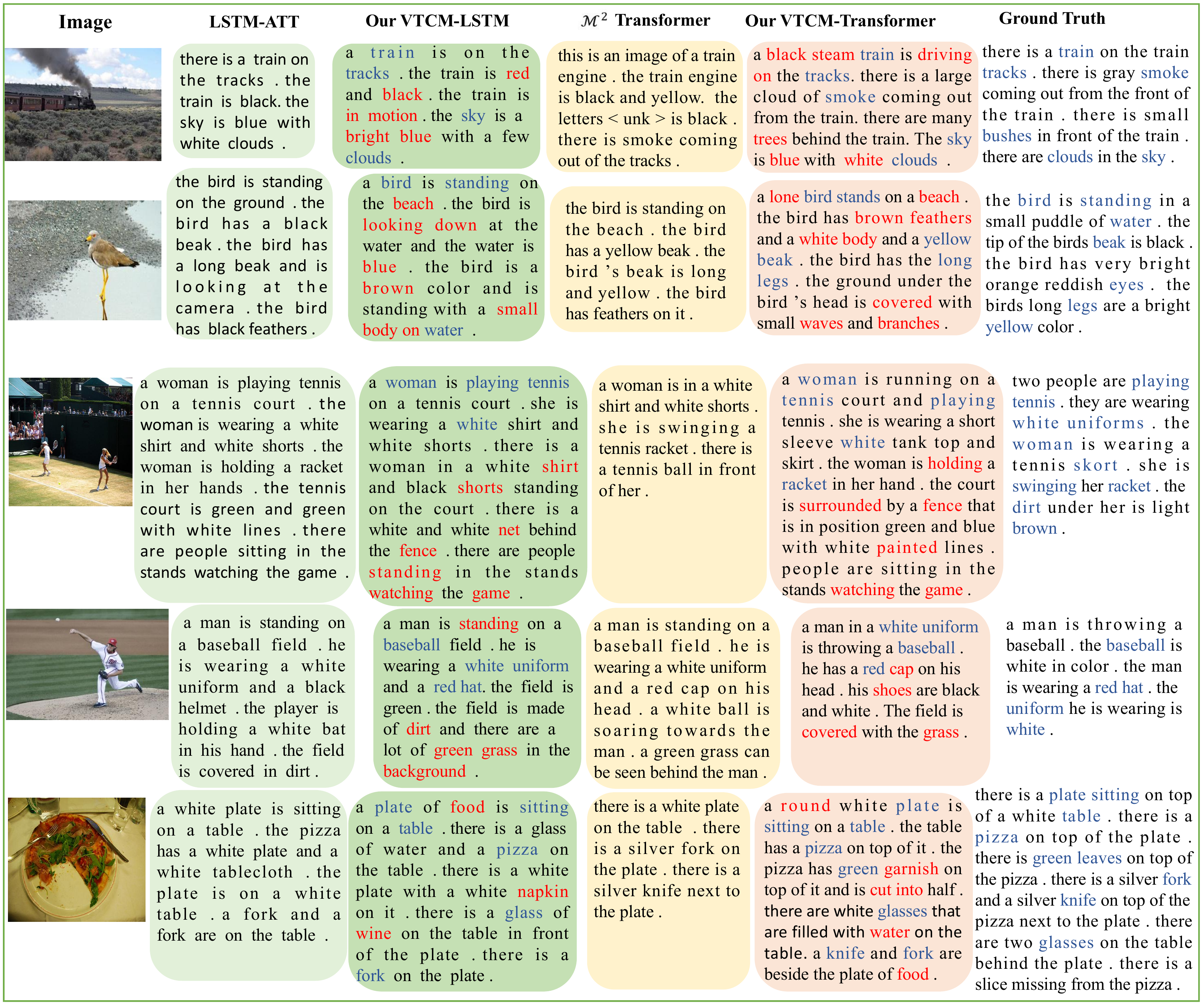}
\caption{{{Examples for paragraphs generated by LSTM-ATT, the proposed VTCM-LSTM, $\mathcal{M}^{2}$ Transformer, the proposed VTCM-Transformer, and human-annotated Ground Truth paragraphs on the Stanford
dataset. (For better visualization, the novel words are colored  in red, the key words of the generated paragraphs and ground truth paragraphs are colored  in blue.)}}}
\label{fig:generation}
\end{center}
\end{figure*}
\noindent\textbf{Ablation Study}:
Firstly, we investigate the impact of topic layers on captioning performance. As it can be seen in Table {\ref{ablation}}, our proposed VTCM-LSTM {and VTCM-Transformer} can produce the desired improvement as its number of layers increases, showcasing the benefits of extracting multi-layer semantic topics for generating descriptive paragraphs. Secondly, to compare our proposed VTCM with other topic models, we adopt PGBN and two representative NTMs as variants to adapt the image paragraph generation task, where LDA with product of experts (ProdLDA) of \citealt{ProdLDA} presents the effective VAE-based inference algorithm for LDA and uses logistic normal distribution for the Dirichlet prior and Dirichlet VAE (DVAE) of \citealt{DVAE} introduces a novel method based on rejection sampling variational inference. Specifically, for PTM, we first train PGBN on the training captions and learn a downstream topic estimator over image features to approximate the topic proportions $\thetav^{1:L}$ in a two-stage manner; for NTMs, including ProdLDA and DVAE, we replace the BoW representation with the visual features as the input into encoder and optimize them with the paragraph generator jointly, where both ProdLDA and DVAE can only extract shallow semantic topics. Although being effective, PGBN-LSTM and PGBN-Transformer are still inferior to their respective opponents using VTCM, which suggests the benefit of introducing the variational topic encoder. The reason behind this might be that the variational topic encoder learns the hierarchical semantic topics by capturing the correlations between image features and descriptive text jointly. Even though the variants of ProdLDA and DVAE can achieve comparable performance with our proposed VTCM-LSTM and VTCM-Transformer at layer $1$, they can not capture the multi-stochastic-layer semantic topics like our VTCM, limiting their ability to generate more coherent paragraphs. Last,to evaluate the effectiveness of our way for integrating the topic information into the LMs, we provide two simple variants of our proposed models, $i.e.$, Topic+LSTM and Topic+Transformer where the topic information is directly concatenated to the output of ahead of the softmax at each time step, based on our adopted hierarchical LSTM and $\mathcal{M}^{2}$-Transformer. The proposed LSTM-based and Transformer-based models both outperform their corresponding base variants, which clearly indicates the usefulness of our proposed ways of incorporating the multi-layer semantic topics into the language decoding process.

\begin{table}[!ht]
{\centering
\caption{{Comparison of the proposed VTCM-LSTM and baselines trained with different strategies.}}
\resizebox{0.5\textwidth}{!}{
\begin{tabular}{c|cccccc}
\toprule[1pt]
\textbf{Method} & METEOR & CIDEr &BLEU-4\\ \hline
\textbf{Our VTCM-LSTM (CE)}&{17.52} &22.82&{9.63}\\
CAE-LSTM (CE+SC) (\citealt{wang2019convolutional}) &18.82 &25.15 &9.67\\
S2TD(CE+SC)(\citealt{shi2021s2td}) &17.64 &24.33 &10.17\\
RAMP (CE+SC)(\citealt{xu2021retrieval})  &17.49 &23.22 &\textbf{10.48}\\
\textbf{Our VTCM-LSTM (CE+SC)}&\textbf{18.95} &\textbf{25.50}&9.88\\
\bottomrule
\end{tabular}}\label{RL_VTCM_LSTM}}
\end{table}

\begin{table}[!ht]
\centering
\caption{{Ablation study on Stanford dataset. Here,  M, C and BN are short for the METEOR, CIDEr and BLEU-N, respectively.}}
\vspace{3mm}
\resizebox{0.5\textwidth}{!}{
\begin{tabular}{c|cccccc}
\toprule[1pt]
\textbf{Method} & M & C &B1 &B2 &B3 &B4\\ \hline
ProdLDA-LSTM&16.22 &18.52 &42.32 &24.86 &15.07 &9.07\\
DVAE-LSTM& 15.77 &18.35 &41.63 &24.47 &14.44 &8.96\\
PGBN-LSTM L=1 &16.15 &18.69 &42.02 &24.83 &15.11 &9.04\\
PGBN-LSTM L=2 &16.37 &19.28 &42.36 &24.80 &15.22 &9.23\\
PGBN-LSTM L=3 &{17.16}&{20.49}&{42.64}&{25.17}&{15.23}&{9.27}\\
\hline
{VTCM-LSTM L=1} &16.50 &19.10 &42.39 &25.42 &15.41 &9.33\\
{VTCM-LSTM L=2} &16.66 &19.98 &42.45 &25.39 &15.46 &9.34\\
\textbf{VTCM-LSTM L=3} &\textbf{17.52}&\textbf{22.82}&\textbf{42.80}&\textbf{25.50}&\textbf{15.69}&\textbf{9.63}\\
\hline
ProdLDA-Transformer &15.56 &21.45 &39.76 &22.36 &14.18 &8.54 \\
DVAE-Transformer &15.21 &21.12 &39.24 &21.87 &13.96 &8.32 \\
PGBN-Transformer L=1 &15.41 &21.35&39.55&22.40&14.17&8.51\\
PGBN-Transformer L=2 &16.18&23.54&40.11&23.25&14.72&9.08\\
PGBN-Transformer L=3 &16.22 &24.83 &40.61 &25.33 &15.51 &9.96\\
\hline
VTCM-Transformer L=1 &15.87 &22.71&39.61&22.92&14.21&8.65\\
VTCM-Transformer L=2 &16.31&23.86&40.17&23.74&15.01&9.16\\
\textbf{VTCM-Transformer L=3} &\textbf{16.88} &\textbf{26.15} &\textbf{40.93} &\textbf{25.51} &\textbf{15.94} &\textbf{9.96}\\
\hline
{Topic+LSTM L=3}& 15.47&18.02&41.80&24.61&14.74&9.10\\
{Topic+Transformer L=3} &15.66&23.45&38.77&23.14&14.51&8.87\\
\bottomrule[1pt]
\end{tabular}}
\label{ablation}
\end{table}

\subsection{Qualitative Evaluation}
\textbf{Generated Captions}:
To qualitatively show the effectiveness of our proposed methods, we show descriptions of different images generated by different methods in Fig. \ref{fig:generation}. As we can see, all of these models can produce paragraphs related to the given images, while our proposed VTCM-LSTM and VTCM-Transformer can generate more coherent and accurate paragraphs by learning to distill the semantic topics from an image via the VTCM module to guide paragraph generation. Therefore, instead of only attending to some visually salient image regions, the generated descriptions of our models are also highly related to the given images in terms of their semantic meanings but not necessarily the words same as the original caption. Taking the first row as the example, the proposed VTCM-Transformer can generate coherent and meaningful paragraphs to describe the image, while capturing meta-concepts like ``steam train'' and ``driving on'' based on the scenes including ``train'' and ``smoke''. Notably, these concepts are even not described in the ground truth but are very relative to the whole image. However, without the high-level semantic information, these baselines  tend to only describe all the salient visual objects in the image, ignoring the ``main plot'' underlying the images, such as the ``the train is in motion'' in the first row and ``food'' in the last row in Fig. \ref{fig:generation}. These observations suggest that the proposed VTCM has successfully captured hierarchical semantic topics by matching visual features to descriptive texts with a similar VAE structure, and our proposed ways of assimilating the topic information into LSTM or Transformer can successfully guide the paragraph generation.
\\
\begin{figure}
\begin{center}
\includegraphics[height=6.8cm,width=8.5cm]{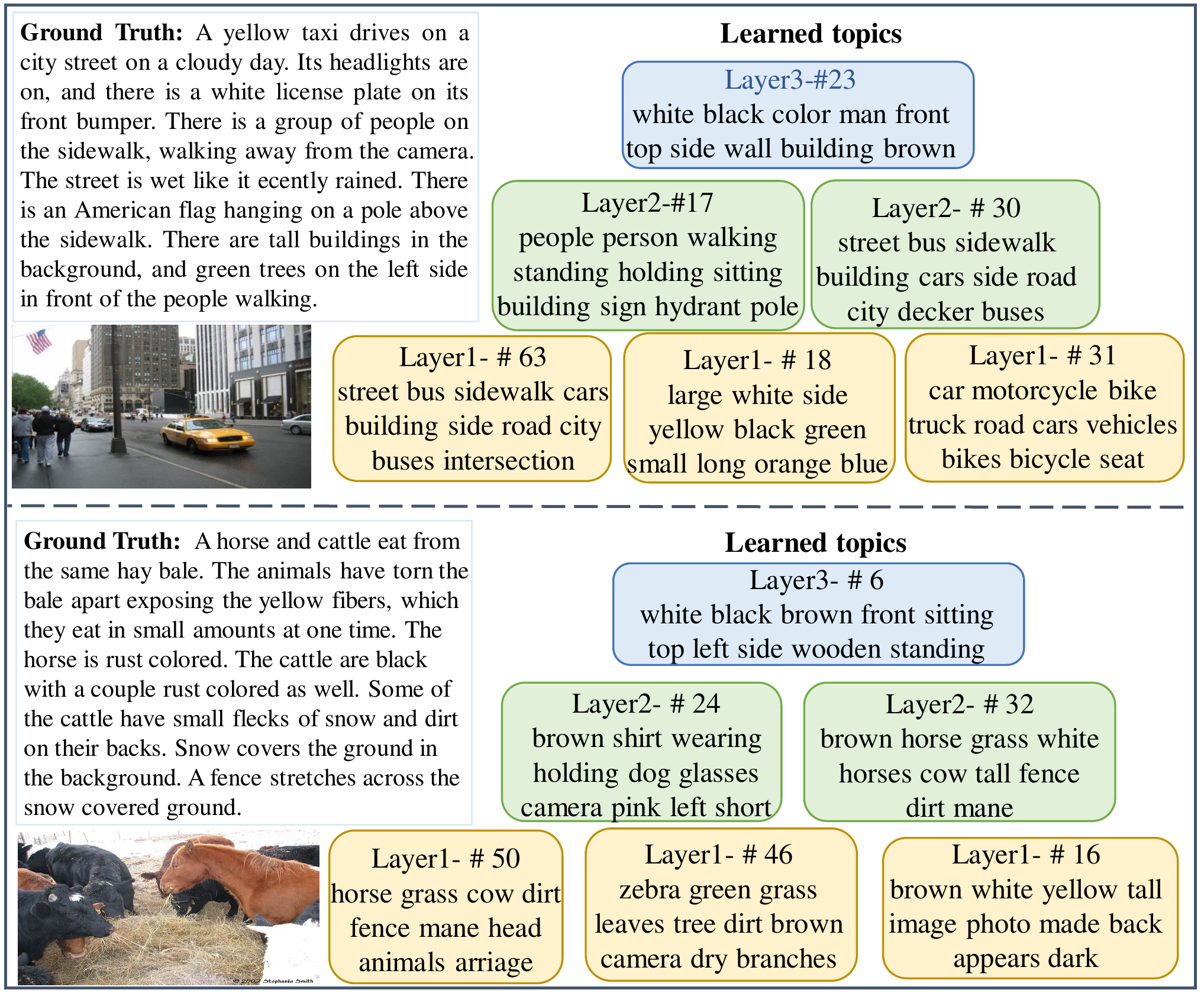}
\caption{{Visualization of the learned topics given the test images, where the top words of each topic at layers 3, 2, and 1 are shown in blue, green, and yellow boxes, respectively. We also present the corresponding ground truth caption for each image, which is not visible at the testing stage.}}\label{fig:learned_topics}
\end{center}
\end{figure}

\begin{figure*}[!htbt]
\begin{center}
\includegraphics[height=5.5cm,width=15cm]{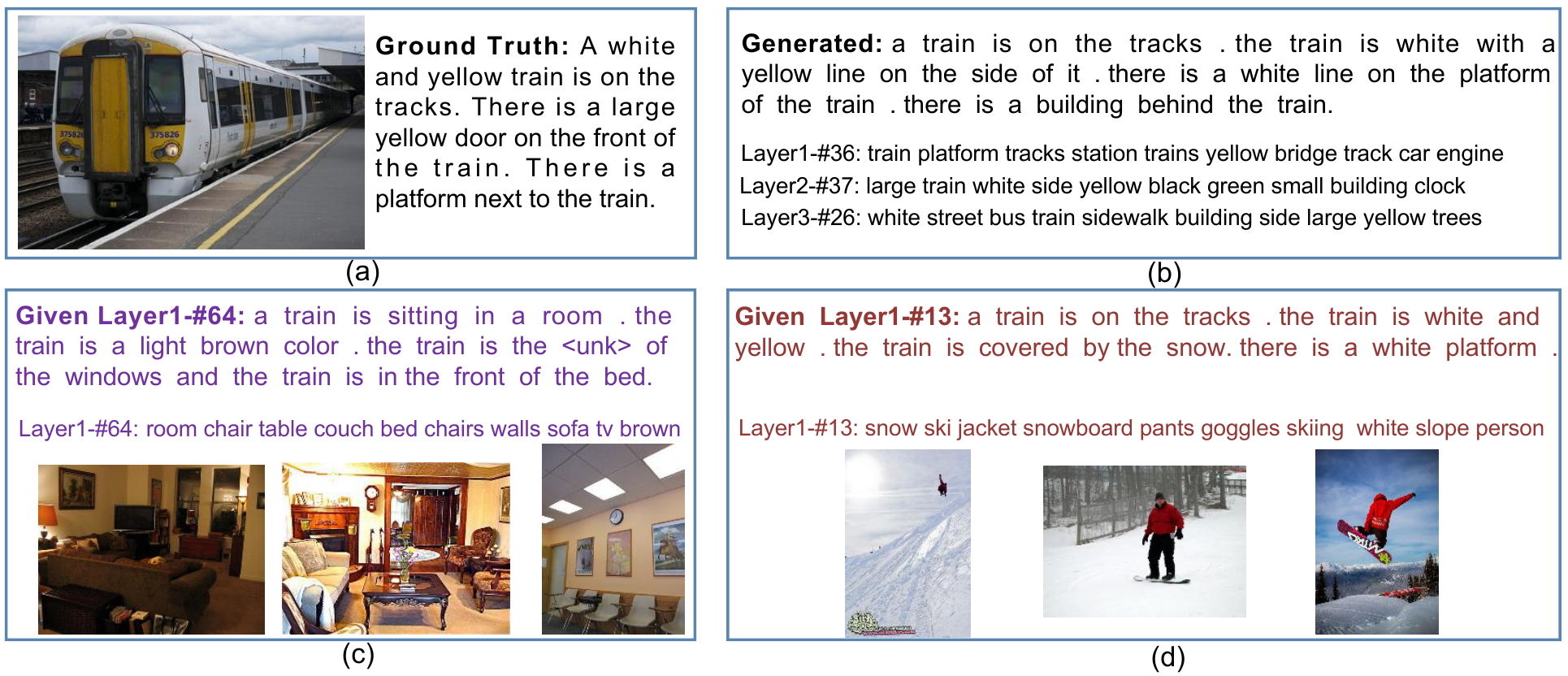}
\caption{{(a) Ground truth caption of an image from Stanford image-paragraph dataset (id = 2349394). (b) The caption generated by the proposed VTCM-LSTM and the corresponding hierarchical topics of the image. The generated caption is able to properly describes the image with the guidance of the correct topic information for paragraph generation. (c)-(d) The generated captions under the guidance of our selected topics. To see the influence of topic information, we replace the original topics with other randomly-selected topic vectors \#64 and \#13 at layer 1. It can be seen that the generated captions are influenced by both the input image and the given topics.
} }
\label{fig:distort}
\end{center}
\end{figure*}

\noindent\textbf{Learned Topics}: One of the benefits of introducing {the hierarchical semantic topics} learned from VTCM is the enhancement of model interpretability. To examine whether the topic model can learn the desired topics from the input image, in Fig. \ref{fig:learned_topics}, we visualize the learned hierarchical topics with our topic model given the input images from the test set, where each topic in different layers has a list of representative words with decreasing ranks. It is clear that the extracted multi-layer topics are highly correlated with the chosen image and its corresponding text. In other words, the topic model we use successfully capture the semantically related topics given the image features. {However, the interpretability of semantic topics is not possessed by the topics in most existing image paragraph captioning models, which are computed by RNN or CAE.} Besides, we can see that the topics become more and more specific when moving from the top layer to the bottom layer. Furthermore, we note that the same scene from different topics, such as the word ``brown,'' is described from multiple perspectives, {making it possible to describe the input image from different topic perspectives. Under the guidance of visual features and the corresponding interpretable hierarchical topics, our model can thus produce a more relevant description for the image.}

\noindent\textbf{Effect of Topics on Paragraph Generation}: We hypothesize that the topic information learned from the image visual features can guide the language paragraph generation model to describe the images. Benefiting from the interpretable semantic topics, our model supports the personalized paragraph generation by manipulating the topic information fed into the LSTM-based or Transformer-based LMs. As shown in Fig. \ref{fig:distort}, for the same image describing a train on the tracks, we make a comparison between the generated captions conditioned on the correct topics predicted by VTCM and the distorted topics. Specifically, the proposed VTCM can infer the image's topic proportion at layer $l$ over global topic $\Phimat^{l} \in \mathbb{R}_+^{K_{l-1} \times K_{l}}$ as $\thetav^{l} \in \mathbb{R}_+^{K_{l}}$, which weights the importance of the $K_{l}$ topics. Therefore, by choosing the index with a maximum value from $\thetav^{l}$ at each layer, we can identify the image's most related topics as \#36, \#37 and \#26 at layers 1, 2 and 3, respectively. Since every topic is a list of representative words, it is easy for users to revise or specify the topic information fed into the paragraph generator by only changing the topic proportion over global topics, where we refer to the designated topic proportion as $\hat{\thetav}^{l}$. For example, to specify the $k$-th topic at layer 1 for the testing image, we can build the one-hot vector $\hat{\thetav}^{1}$ as the distorted topic proportion, where $\hat{\theta}^{1}_{k'}=1$ only if ${k'}=k$. And we set the $\hat{\thetav}^{2:L}$ as zero vectors for simplicity. Clearly, topic \#64 at layer 1 is about the room, and the caption about the image is now changed to regarding the train sitting on the room when we set $\hat{\theta}^{1}_{64}=1$ and feed $\hat{\thetav}^{1:L}$ into the paragraph generator. Similarly, topic \#13 at layer 1 corresponds to the snow and ski, and the caption now is changed to the train covered by the snow.
These observations suggest that the generated captions not only are related to the input image but also reflect what the user wants to emphasize by only changing the topic proportion, making our proposed model controllable.
\begin{figure*}[!htbt]
\begin{center}
\includegraphics[height=8cm,width=17cm]{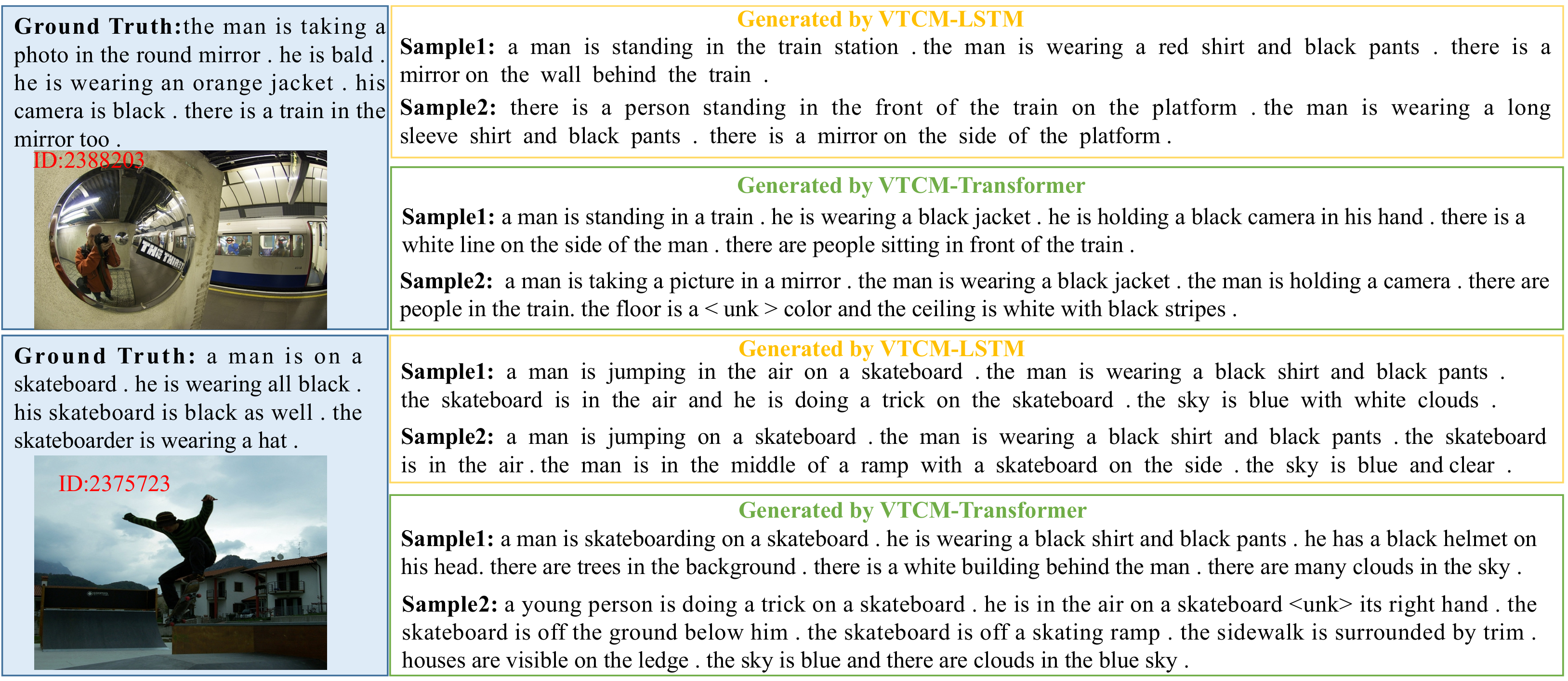}
\caption{Different paragraphs generated by the proposed VTCM-LSTM and VTCM-Transformer for two example images from the Stanford image-paragraph dataset (id = 2388203 and 2375723). {Given the extracted hierarchical topics ${\Phimat^{1:L}}$ with VTCM from each image, we only sample two different uniform noises ${\epsilonv^{1:L}}$ and produce the corresponding topic weight vectors ${\thetav^{1:L}}$ with randomness using Equation \eqref{sample_theta}, which have similar semantic information and are fed into the language model to depict the input image.
}}
\label{fig:diverse}
\end{center}
\end{figure*}

\noindent\textbf{Diversity}:  To show the uncertainty in VTCM makes it capable of producing diverse captions while keeping the ``main plot'' unchanged, we example two descriptions with the same set of inputs, respectively. Different from the bottom row in Fig. \ref{fig:distort}, here we  do not distort the predicted topic proportions from the test image (id=2388203 or id=2375723) but only sample different uniform noises ${\epsilonv^{1:L}}$ to generate $\thetav^{1:L}$ via Equation~\eqref{sample_theta}, whose posterior parameters transformed from the image features are kept unchanged. Therefore, different noises ${\epsilonv^{1:L}}$ lead to  different random topic weight vectors $\thetav^{1:L}$, which however share similar weights over global topics and thus similar semantic information. As shown in Fig. \ref{fig:diverse}, our proposed models can generate diverse and coherent paragraphs while ensuring the ``big picture'' underlying the image does not get lost in the details. The reason behind this might be that our frameworks feed the multi-stochastic-layer latent topic representation $\thetav^{1:L}$ of VTCM as the source of randomness to the language generator. Benefiting from the assimilation of multi-stochastic-layer topic information into the language generator, our proposed topic-guided image paragraph captioning systems can guarantee diversity and produce diversified outputs even if there is no specialized module.
\begin{figure}
\begin{center}
\includegraphics[height=5cm,width=8.5cm]{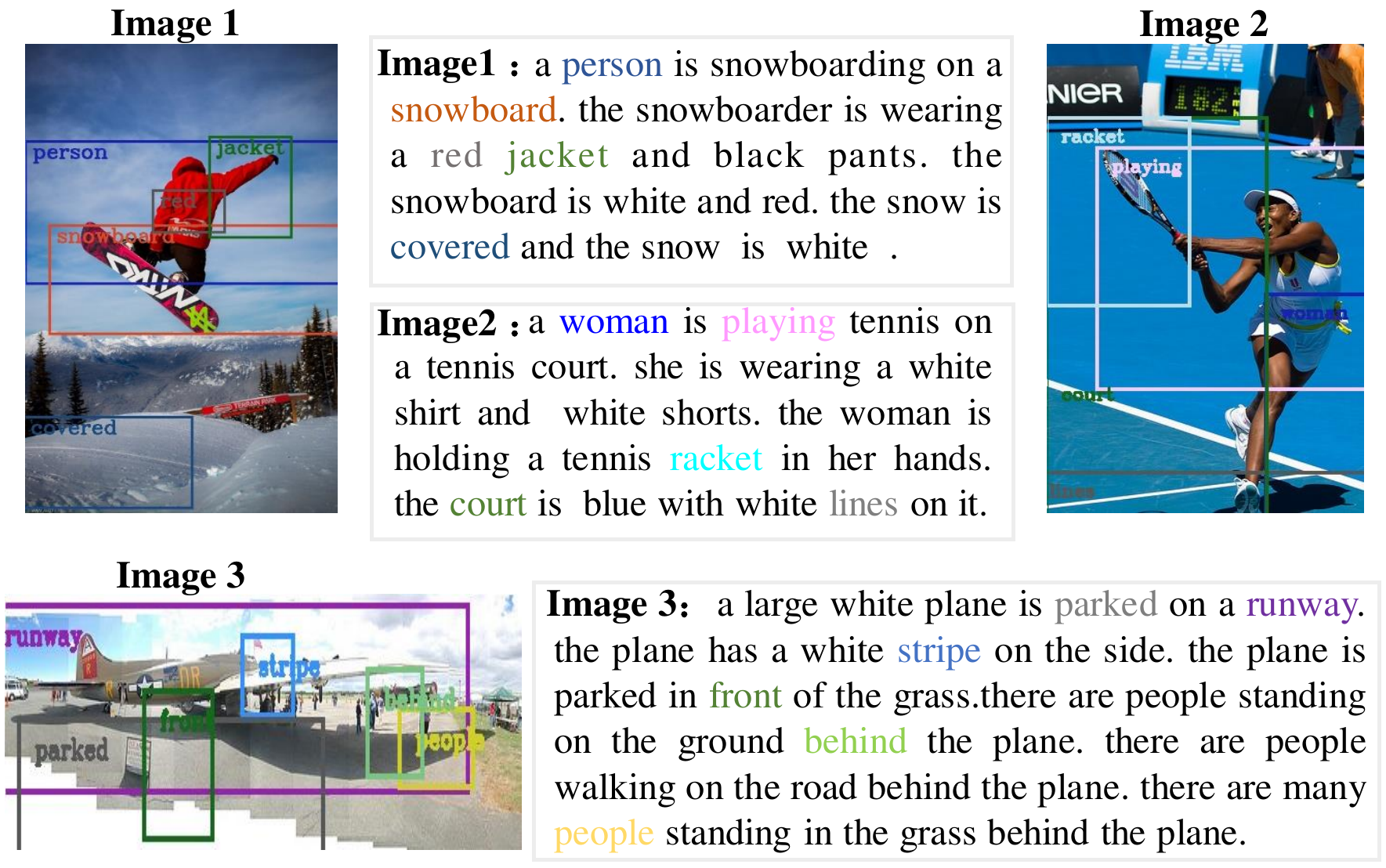}
\caption{{ Example of generated captions by VTCM-LSTM showing attended image regions. Different regions and the corresponding words are shown in the same color. }}
\label{fig:attention}
\end{center}
\end{figure}
\\
\textbf{The Attention Mechanism in VTCM-LSTM}: To evaluate the effectiveness of the attention mechanism in VTCM-LSTM on image captioning, in Fig. \ref{fig:attention}, we visualize the attended image regions with the biggest attention weight for different words. As we can see, our proposed VTCM-LSTM can reason where the model is focusing on at different time steps. Here we take Image 1 as an example. When predicting ``person,'' the attention module precisely chooses the bounding box covering the main part of the body. While predicting the word ``snowboard,'' our model decides to attend to the surrounding area about the snowboard. It proves that our proposed VTCM-LSTM can capture the alignment between the attended area and the predicted word, which reflects the human intuition during object description.

\section{Conclusion}
We develop a plug-and-play hierarchical-topic-guided image paragraph generation pipeline, which couples a visual extractor with a deep topic model to guide the learning of a language paragraph generation model. As a visual-textual coupling model, the deep topic model can capture the correlations between the image and text at multiple levels of abstraction and learn the semantic topics from images. Serving as the bridge between two modalities, the distilled hierarchical topics are  used to guide the caption generation in language model, where we remould both the LSTM-based and Transformer-based language models. Experimental results on the Stanford paragraph dataset show that our proposed models outperform a variety of competing paragraph captioning models, while inferring interpretable hierarchical latent topics and generating  semantically coherent paragraphs for the given images.

\begin{acknowledgements}
B. Chen acknowledges the support of NSFC (U21B2006 and 61771361), Shaanxi Youth Innovation Team Project, the 111 Project (No. B18039) and the Program for Oversea Talent by Chinese Central Government. M. Zhou acknowledges the support of 
NSF IIS-1812699.
\end{acknowledgements}



\appendix
\section{Appendix}
\subsection{The Variational Topic Encoder of VTCM}\label{sec:topic encoder}
Inspired by \citealt{zhang2018whai}, to approximate the gamma distributed topic weight vector $\thetav^{l}$ with a Weibull distribution, we assume the topic encoder as
$q(\thetav^{l}|\overline{\vv}) = \mbox{Weibull}(\kv^{l},\lambdav^{l})$, where the parameters $\kv^{l}$ and $\lambdav^{l}$ of $\thetav^{l}$ can be denoted as
 \begin{align}
&\kv^{l} = \ln[1+\exp(\Wmat_{hk}^{l}\hv^{l}+ \bv_1^{l})] , \label{MLP1} \\
&\lambdav^{l} = \ln[1+\exp(\Wmat_{h \lambda}^{l}\hv^{l}+ \bv_2^{l})] , \label{MLP2}
\end{align}
where $\hv^{l}$ are deterministically nonlinearly transformed from the image pooled representation, stated as $\hv^{0}=\overline{\vv}$ and $ \hv^{l}\!=\!\mbox{tanh} \left(\Wmat_v^{l}\hv^{l-1}+\bv_v^{l}\right)$.

\subsection{The Gating Unit in VTCM-LSTM}\label{sec:GRU_gate}
Note the input $\uv_{j,t}^{l}$ of sentence-level LSTM at layer $l$ combines the topic weight vectors $\thetav^{l}$ and hidden output of the sentence-level LSTM $\hv_{j,t}^{s,l}$ at each time step $t$. To realize $\uv_{j,t}^{l}=g\left(\hv_{j,t}^{s,l}, \thetav^{l}\right)$, we adopt a gating unit similar to the gated recurrent unit (GRU) (\citealt{RNNencoder-decoder}), defined as
\begin{align}\label{coupling_vector}
\uv_{j,t}^{l}  &=\left(1-\zv_{j,t}^{l}\right) \odot \hv_{j,t}^{s,l} + \zv_{j,t}^{l} \odot \hat{\hv}_{j,t}^{s,l}\,\, .
\end{align}
where
\begin{align}\label{coupling_vector}
\zv_{j,t}^{l}   &= \sigma\left(\Wmat_{z}^{l}  \thetav^{l} + \Umat_{z}^{l}  \hv_{j,t}^{s,l}  +  \bv_{z}^{l} \right), \nonumber\\
\rv_{j,t}^{l} & = \sigma\left(\Wmat_{r}^{l} \thetav^{l}  + \Umat_{r}^{l}  \hv_{j,t}^{s,l} + \bv_{r}^{l} \right), \\
\hat{\hv}_{j,t}^{s,l} & =\tanh \left(\Wmat_{h}^{l} \thetav^{l} + \Umat_{h}^{l}\left(\rv_{j,t}^{l} \odot \hv_{j, t}^{s,l}\right)+\bv_{h}^{l}\right). \nonumber
\end{align}
Define $\uv_{j,t}^{1:L}$ as the concatenation of $\uv_{j,t}^{l}$ across all layers and $\Wmat_o$ as a weight matrix with $V$ rows, the conditional distribution probability $p\left(w_{j,t} \given w_{j,<t}, \emph{Img}\right)$ of $w_{j,t}$ becomes
\begin{equation}\label{word_LM}
 p\left(w_{j,t} \given w_{j,<t}, \vv_{1:M},\thetav^{1:L}\right) =  \operatorname{softmax}\left(\Wmat_o \uv_{j,t}^{1:L}\right).
 \end{equation}
 There are two advantages to combine $\uv_{j,t}^{l}$ at all layers for language generation. First, the combination can enhance representation power because of different statistical properties at different stochastic layers of the deep topic model. Second, owing ``skip connections'' from all hidden layers to the output, one can reduce the number of processing steps between the bottom of the network and the top, mitigating the ``vanishing gradient'' problem (\citealt{alex2013}).

\subsection{Likelihood and Inference of VTCM-Transformer}\label{sec:likelihood_VTCM_Transformer}
Given an image $\emph{Img}$, we can also represent the paragraph as $\boldsymbol{Y}=\{y_1,...,y_I\}$, which is suitable for flat language model, such as Transformer-based model. Under the deep topic model (VTCM) and Transformer-based LM, the joint likelihood of the target ground truth paragraph $\boldsymbol{Y}$ of $\emph{Img}$ and its corresponding BoW count vector $\dv$ is defined as
\begin{align}\label{likelihoodTrans}
 &\resizebox{0.99\hsize}{!}{$p\left( \boldsymbol{Y},\dv \given \Phimat^{1:L},\vv_{1:M}\right)\!=\!
 \int p\left(\dv\given \Phimat^{1}\thetav^{1}\right) \left[\prod_{l = 1}^{L} p\left(\thetav^{l}\given \Phimat^{l+1}\thetav^{l+1}\right) \right]$}\notag\\
&\resizebox{0.65\hsize}{!}{$ \prod_{i=1}^{I}   p\left(y_{i} \given y_{<i}, \vv_{1:M},\thetav^{1:L}\right)
d{\thetav^{1:L}}$},
\end{align}
Since we introduce a variational topic encoder to learn the multi-layer topic weight vectors $\thetav^{1:L}$ with the image features $\overline{\vv}$ as the input. Thus, a lower bound of the log of \eqref{likelihoodTrans} can be constructed as
 \begin{align}\label{ELBO-of-Transformer}
 L_\text{all} &= \mathbb{E}_{q(\thetav^{1}|\overline{\vv})}\left[ \ln p\left( \dv \given \, \Phimat^{1}\thetav^{1} \right)\right] \nonumber\\
 &- \sum_{l=1}^L \mathbb{E}_{q(\thetav^{l}|\overline{\vv})} \left[ \ln \frac{q\left( \thetav^{l}\given \overline{\vv}  \right) }{p \left( \thetav^{l} \given
\Phimat^{l + 1}\thetav^{l+ 1}\right)} \right]\\
 &+\sum_{l=1}^L \mathbb{E}_{q(\thetav^{l}|\overline{\vv})} \left[ \sum_{i=1}^{I} \ln p\left( {y_{i}}\given y_{<i},\vv_{1:M},\thetav_j^{1:L}\right)\right],\nonumber
\end{align}
which unites the first two terms primarily responsible for training the topic model component, and the last term for training the Transformer-based LM component. The parameters $\Omegamat_{\text{TM}}$ of the variational topic encoder and the parameters $\Omegamat_\text{Trans}$ of Transformer-based LM can be jointly updated by maximizing  $L_\text{all}$. Besides, the global parameters $\Phimat^{1:L}$ of the topic decoder can be sampled with TLASGR-MCMC in \citet{cong2017deep} and presented below. The training strategy of VTCM-Transformer is similar to that of VTCM-LSTM.

\subsection{Inference of Global Parameters $\Phimat^{1:L}$ of VTCM}\label{sec:update_Phi}
For scale identifiability and ease of inference and interpretation, the Dirichlet prior is placed on each column of $\boldsymbol{\Phi}^{l} \in$ $\mathbb{R}_{+}^{K_{l-1} \times K_{l}}$, which means $0 \leq \Phi^{l}_{k^{\prime}, k} \leq 1 $ and $\sum_{k^{\prime}=1}^{K_{l-1}} \Phi^{l}_{k^{\prime}, k}= 1 $. To allow for scalable inference, we apply the topic-layer-adaptive stochastic gradient Riemannian (TLASGR) MCMC algorithm described in \cite{cong2017deep,zhang2018whai}, which can be used to sample simplex-constrained global parameters  in a mini-batch based manner. It improves its sampling efficiency via the use of the Fisher information matrix (FIM), with adaptive step-sizes for the topics at different layers.
Here, we discuss how to update the global parameters $\{\Phimat^{l}\}_{l=1}^L$ of VTCM in detail and give a complete one in Algorithm 1.
\\
{{\bf{Sample the auxiliary counts:}}
This step is about the ``upward'' pass. For the given mini-batch $\{ \emph{Img}_n, P_n, \dv_n \}_{n=1}^{N}$ in the training set, $\dv_n$ is the bag of words (BoW) count vector of paragraph $P_n$ for input image $\emph{Img}_n$ and $\thetav_n^{1:L}$ denotes the latent features of the $n$th image. By transforming standard uniform noises ${\epsilonv_n^{l}}$, we can sample $\thetav_n^{l}$ as
\begin{align}\label{sample_theta1}
 \thetav_n^{l} & \small = {\lambdav_n^{l}} \left(-\ln(1-{\epsilonv_n^{l}})\right) ^ {1/{\kv_n^{l}}}.
\end{align}
Working upward for $l = 1,...,L$, we can propagate the
latent counts $x_{vn}^{l}$ of layer $l$ upward to layer $l + 1$ as
\begin{align}\label{Multi_Phi_Theta}
 & \resizebox{0.99\hsize}{!}{$ {A_{v{1:K_l}n}^{l}}\!\sim\! \mbox{Multi}\left(x_{vn}^{(l)};\!\frac{{\phi _{v{1}}^{l}\theta _{{1}n}^{l}}}{{\sum\nolimits_{{k_l} = 1}^{{K_l}} {\phi_{v{k_l}}^{l}\theta _{{k_l}n}^{l}} }},\!...\!,\frac{{\phi _{v{K_l}}^{l}\theta _{{K_l}n}^{l}}}{{\sum\nolimits_{{k_l} = 1}^{{K_l}} {\phi _{v{k_l}}^{l}\theta_{{k_l}n}^{l}} }}\right),$}\\
 &  \mv_{kn}^{(l)(l+1)} = \sum_{v = 1}^{{K_{l-1}}} {A_{vkn}^{ {l}}},\\
 & x_{kn}^{(l+1)}\sim \mbox{CRT}\left(\mv_{kn}^{(l)(l+1)}, \phiv_{k:}^{l+1}\thetav_{n}^{l+1}\right),
\end{align}
where $x_{vn}^{1}=d_{vn}$, $\dv_{n}=\{d_{1n},..,d_{vn},..,d_{{V_c}n}\}$, $V_c$ is the size of vocabulary in VTCM, and $x_{kn}^{(l+1)}$ denotes the latent counts at layer $l+1$.
\\
{\bf{Sample the hierarchical components $\{\Phimat^{l}\}_{l=1}^L$: }}For $\phiv_k^{l}$, the $k$th column of the loading matrix $\Phimat^{l}$ of layer $l$, its sampling can be efficiently realized as
\begin{align}\label{TLASGR update_Phi}
\left( {\phiv_k^{l}} \right)_{q + 1} \! = & \! \bigg[  \resizebox{0.8\hsize}{!}{$ \! \left( {\phiv_k^{l}} \right)_q \! + \! \frac{\varepsilon _q}{P_k^{l}} \! \left[ \left(\rho \tilde \Av_{:k\cdotv}^{l} \! + \! \eta_{0}^{l}\right) \! - \! \left(\rho \tilde A_{\cdotv k\cdotv}^{l} \! + \! K_{l-1}\eta_{0}^{l} \right) \! \left( {\phiv_k^{l}} \right)_q  \right] $}
\! \nonumber \\
&+\!  \mathcal{N} \left( 0, \frac{2 \varepsilon _n}{P_k^{l}}\left[ \mbox{diag}(\phiv_k^{l})_q - (\phiv_k^{l})_q (\phiv_k^{l})_q^T \right] \right) \bigg]_\angle,
\end{align}
where $\varepsilon_q$ denotes the learning rate at the $q$th iteration, $\rho$ the ratio of the dataset size to the mini-batch size, $P_k^{l}$ is calculated using the estimated FIM, $\tilde A_{{k'}k\cdotv}^{{l}} = \sum_{{n} = 1}^{N} {A_{{k'}kn}^{ {l}}}, {\tilde \Av_{:k\cdotv}^{l }} = \{\tilde A_{{1}k\cdotv}^{{l}},\cdots, \tilde A_{{K'}k\cdotv}^{{l}} \}^{T} $ and ${\tilde A_{\cdotv k\cdotv}^{l}}= \sum_{{k'} = 1}^{K'} \tilde A_{{k'}k\cdotv}^{{l}} $, ${A_{{k'}kn}^{ {l}}}$ comes from the augmented latent
counts $A^{l}$ in \eqref{Multi_Phi_Theta},
$\eta_{0}^{l}$ is the prior of ${\phiv_k^{l}}$,
and $[\cdot]_\angle$ denotes a simplex constraint. More details about TLASGR-MCMC for our proposed model can be found in the Equations (18-19) of \cite{cong2017deep}.
\begin{figure*}[!t]
\begin{center}
\includegraphics[height=5.9cm,width=12.2cm]{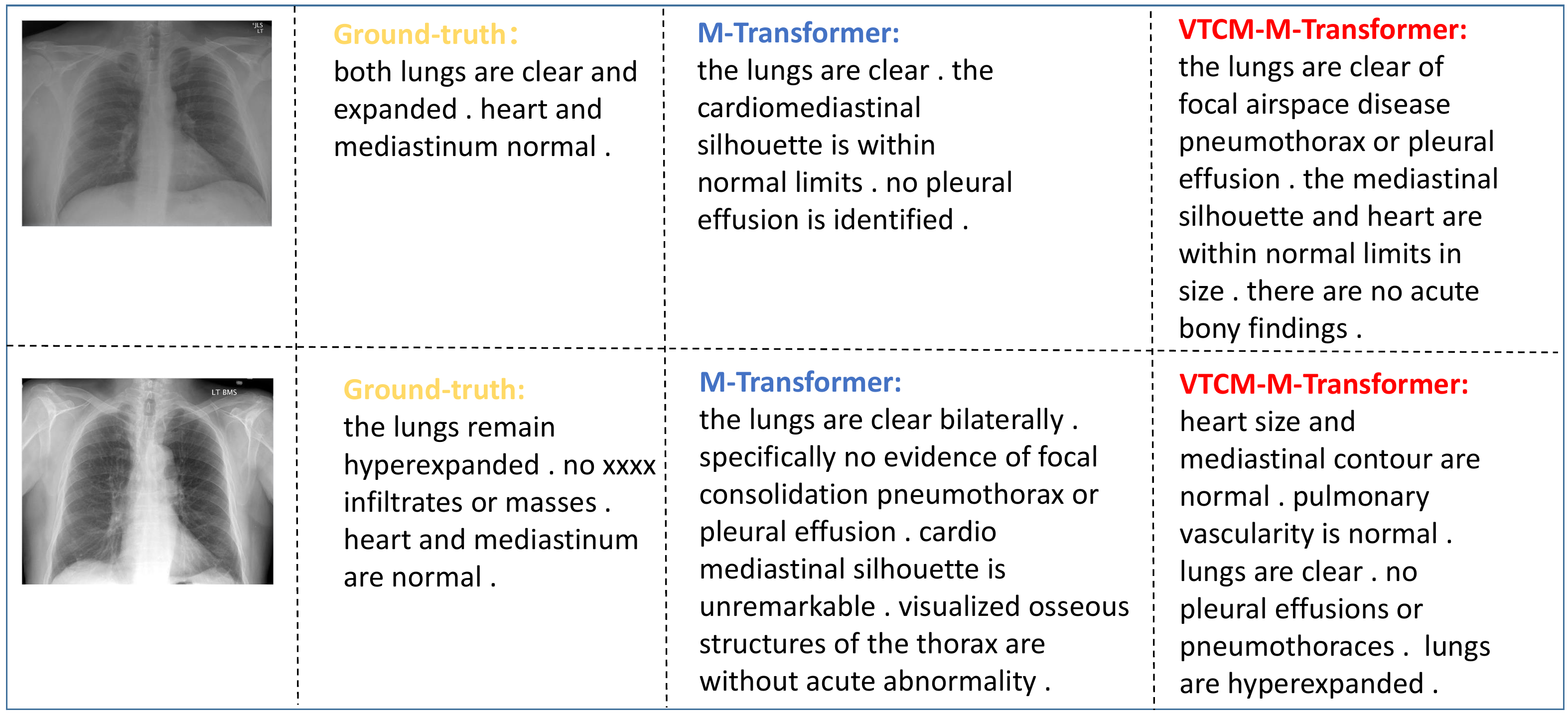}
\caption{Illustrations of reports from ground-truth, M-Transformer and VTCM-M-Transformer models for two X-ray chest images.}
\label{fig:transformer}
\end{center}
\end{figure*}

\begin{table*}[!ht]
\centering
\caption{Comparisons of our proposed VTCM-M-Transformer with M-Transformer on the test sets of IU X-RAY and MIMIC-CXR, where RG-L is ROUGE-L.}
\resizebox{0.8\textwidth}{!}{
\begin{tabular}{c|c|cccccc}
\toprule[1pt]
\textbf{Data}&\textbf{Method} &B-1 &B-2 &B-3 &B-4 & M & RG-L\\ \hline
\multirow{2}{0.8in}{IUX-RAY}& {M-Transformer}     &0.470  &0.304  &0.219  &\textbf{0.165} &0.187 &0.371  \\
&{\textbf{VTCM-M-Transformer}} &\textbf{0.495} &\textbf{0.314} &\textbf{0.222} &0.162 &\textbf{0.191} &\textbf{0.376} \\
\hline
\multirow{2}{0.8in}{{MIMIC-CXR}}&{M-Transformer} &0.353 &0.218  &0.145 &0.103 &0.142& 0.277\\
&{\textbf{VTCM-M-Transformer}} & \textbf{0.367} & \textbf{0.224} &\textbf{0.150} &\textbf{0.110} &\textbf{0.149} &\textbf{0.286}\\
\bottomrule
\end{tabular}}
\label{medical_report}
\end{table*}

\subsection{Additional Experimental Results}\label{sec:additional_results}
To validate the generalizability of our proposed model, we also conducted experiments on the task of generating the radiology reports for the chest X-ray images, which is an important task to apply artificial intelligence to the medical domain. We consider the memory-driven Transformer (M-Transformer) designed for the radiology report generation task (\citealt{chen2020generating}) as our baseline, which introduces a relational memory (RM) to record the information from previous generation processes and a memory-driven conditional layer normalization (MCLN) to incorporate the memory into Transformer. For a fair comparison, we adopt the same implementation for our model; see  \citet{chen2020generating} for more details. Our experiments are performed on two prevailing radiology report datasets. \textbf{IUX-RAY} (\citealt{demner2016preparing}) is collected by the Indiana University and consists of 7,471 chest X-ray images and 3,955 reports; \textbf{MIMIC-CXR} (\citealt{johnson2019mimic}) includes 473,057 chest X-ray images and 206,563 reports from 63,478 patients. Following \citet{chen2020generating} , we exclude the samples without reports. Note that we can flexibly select the language model for our plug-and-play system, since we pay more attention to assimilating the multi-layer semantic
topic weight vectors into the paragraph generator. We here adopt the same Transformer encoder with the M-Transformer and introduce the three-layer semantic topics into its Transformer decoder, where we only add the concatenated topic proportion $\thetav^{1:L}$ to the embedding vector of the input token $y_{t-1}$, {which are then embedded to calculate the keys $ \boldsymbol{K}$ and values $ \boldsymbol{V}$ of the decoder}. As summarized in Table \ref{medical_report}, our model (VTCM-M-Transformer) outperforms the M-Transformer on
METEOR, ROUGE-L, BLEU-1, BLEU-2 and BLEU-3 and is competitive on the BLEU-4. It indicates that the multi-layer semantic topics with VTCM can enhance the radiology report generation despite without designing the complex language model on purpose, proving the generalizability and flexibility of our model. To qualitatively show the effectiveness of our proposed method, we show the reports of two randomly sampled X-ray chest images generated by different methods in Fig. \ref{fig:transformer}, as well as the ground truth reports. Compared with M-Transformer, our VTCM-M-Transformer produces more detailed and coherent reports. For the first image, our VTCM-M-Transformer describes the ``heart'' and ``mediastinal silhouette'' in a natural way, while the M-Transformer ignores the ``heart''. For the second image, the report generated by our model is closer to the ground-truth report, which summarizes the ``lungs are hyperrexpanded''. The above observations show that using the hierarchical semantic topics
can enhance the radiology report generation.

\bibliographystyle{spbasic}
\bibliography{ImagePara}
\end{document}